\documentclass{article}

\usepackage{ember_article,times}

\usepackage{amsmath,amssymb}
\usepackage{graphicx}

\usepackage{float}
\usepackage{booktabs}
\usepackage{hyperref}
\usepackage{url}
\usepackage{xcolor}
\usepackage{colortbl}
\usepackage{tcolorbox}
\definecolor{best}{HTML}{D9ECE0}
\usepackage{multirow}%
\usepackage{array,tabularx}
\definecolor{urlteal}{HTML}{55B8B0}
\newcolumntype{L}[1]{>{\raggedright\arraybackslash}p{#1}}
\newcolumntype{C}[1]{>{\centering\arraybackslash}p{#1}}
\newcolumntype{Y}{>{\raggedright\arraybackslash}X}
\newtcolorbox{promptbox}{
    colback=black!5,
    colframe=black!20,
    boxrule=0.4pt,
    arc=2pt,
    boxsep=0pt,
    left=8pt,
    right=8pt,
    top=6pt,
    bottom=6pt,
    fontupper=\normalfont\normalsize,
    before upper=\raggedright
}

\newcommand{\rwWork}[2]{\textbf{#1}~{\scriptsize\citeyearpar{#2}}}
\newcommand{\rwOwn}{\textbf{EMBER-Bench (Ours)}}

\newcommand{\Y}{\(\checkmark\)}
\newcommand{\benchPartial}{\ensuremath{\triangle}}
\newcommand{\N}{--}

\title{EMBER-Bench: Benchmarking Cross-Event Causal Memory in Long-Horizon Embodied Tasks}

\author{%
\makebox[\textwidth][c]{%
\begin{tabular}{@{}c@{}}
\textbf{Aoyang Cai}$^{1,3,*}$ \quad
\textbf{Boning Zhao}$^{2,3,*}$ \quad
\textbf{Shaoxuan Xie}$^{3}$ \quad
\textbf{Dahui Gao}$^{3}$
\\[0.6pt]
\textbf{Huan Yang}$^{3}$ \quad
\textbf{Zhongyuan Wang}$^{3}$ \quad
\textbf{Zhiwei Yu}$^{3,\dagger}$ \quad
\textbf{Guocai Yao}$^{3,\dagger}$
\\[4pt]
$^{1}$Tsinghua University
\qquad
$^{2}$The University of Hong Kong
\\[0.5pt]
$^{3}$Beijing Academy of Artificial Intelligence
\\[3pt]
{\small
$^{*}$Equal contribution
\qquad
$^{\dagger}$Corresponding authors
}
\\[5pt]
{\small
\textbf{Project Page: }
\url{https://zhaoalexgoat.github.io/EMBER-Bench/}
}
\end{tabular}%
}%
}

\hypersetup{hidelinks}

\usepackage{tcolorbox}
\tcbuselibrary{skins,breakable}

\tcbset{emberpromptstyle/.style={
    enhanced,
    breakable,
    colback=black!5,
    colframe=black!20,
    colbacktitle=black!5,
    coltitle=black,
    boxrule=0.4pt,
    arc=2pt,
    boxsep=0pt,
    left=8pt,
    right=8pt,
    top=6pt,
    bottom=6pt,
    toptitle=6pt,
    bottomtitle=3pt,
    titlerule=0pt,
    fontupper=\normalfont\normalsize,
    fonttitle=\normalfont\normalsize\bfseries,
    before upper={\raggedright\setlength{\parindent}{0pt}\setlength{\parskip}{0pt}},
    before skip=8pt,
    after skip=10pt
}}

\newtcolorbox[auto counter,number within=section]{emberprompt}[2][]{
    emberpromptstyle,
    title={Prompt~\thetcbcounter: #2},
    title after break={Prompt~\thetcbcounter: #2 (continued)},
    #1
}

\newtcolorbox{emberpromptcontext}[1]{
    emberpromptstyle,
    title={#1},
    title after break={#1 (continued)}
}

\begin{document}
\maketitle

\pagestyle{plain}
\thispagestyle{plain}

\begin{abstract}
Lifelong physical agents must reason over extended interactions where past events continue to shape the world long after they disappear from view. Beyond recalling what happened, agents must infer how history changes the current state and constrains future actions. Yet existing embodied and video-memory benchmarks largely focus on historical retrieval and summary, leaving such history-dependent causal reasoning underexplored. We introduce \textbf{EMBER-Bench}, an egocentric benchmark for cross-event causal reasoning in long-horizon embodied tasks, for which we newly created the task design, video recording, and data annotation. It contains 189 household tasks and 699 QA pairs, spanning task progress, failure recovery, external interventions, and compound long-horizon tasks with distant dependencies and prerequisites, with fine-grained event and causal-chain annotations. EMBER-Bench evaluates reasoning in both directions: next-action prediction selects the next action from history, and causal traceback, given that action, identifies the historical event that makes it necessary. Input ablations that add action logs or privileged cause-and-consequence annotations to the video indicate which kind of historical information models fail to use. Among the 16 evaluated models, the highest overall accuracy is 61.2\%, compared with a mean of 98.3\% across two human evaluators. At paired decision points, correct traceback is not associated with correct next-action prediction. Adding action logs yields a gain of 1.6 points, whereas cause-and-consequence annotations yield an additional gain of 13.0 points on top of that. These results suggest that extracting causal information from past events and converting it into constraints on current actions remains a key difficulty for long-horizon embodied agents. Project Page: \url{https://zhaoalexgoat.github.io/EMBER-Bench/}.
\end{abstract}

\section{Introduction}
\label{sec:introduction}
\begin{figure}
    \centering
    \includegraphics[width=1\linewidth]{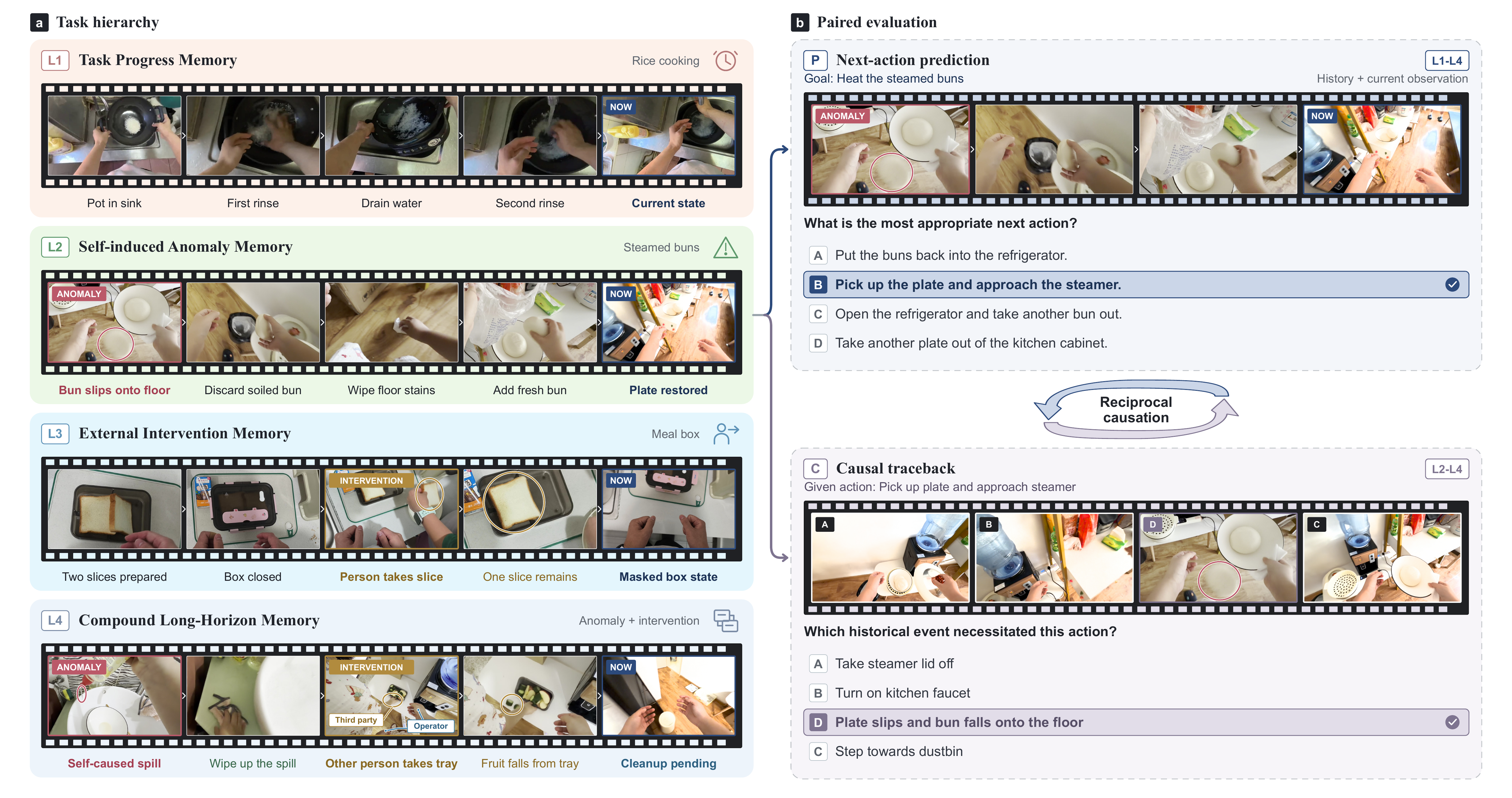}
    \caption{\textbf{Overview of EMBER-Bench.} \textbf{(a)} Four types of memory required for long-horizon task scenarios. \textbf{(b)} Paired next-action prediction and causal traceback assess history-grounded action selection and identification of the historical event supporting a given action.}
    \label{fig:teaser}
\end{figure}
Lifelong agents interacting with the physical world must operate in persistent environments where past events continue to shape later decisions, even across tasks. When the consequences of past events cannot be inferred from the current observation, reliable long-horizon task execution fundamentally requires memory. Recent advances in vision-language-action models (VLAs) offer one route toward this broader vision. $\pi_{0.5}$ \citep{physicalintelligence2025pi05} has demonstrated its ability to perform multi-stage tasks such as kitchen and bedroom tidying in residences not covered by training. Such progress motivates a broader question: when decision-relevant information is no longer visible, can a model use past history to determine which effects persist and choose the appropriate next action?

Task duration alone does not establish memory dependence. When the current scene exposes task progress, or a familiar action sequence suggests the next step, a model may answer correctly without consulting the relevant history. Memory must track the current task state rather than merely retain past observations. This requires distinguishing causally relevant events from incidental history and tracking how their still-valid consequences constrain current actions. We focus on decision points where, given the goal and task constraints, the current observations alone are insufficient to distinguish between task states that require different actions. These requirements include tracking task progress and prerequisites, considering the ongoing consequences of self-induced anomalies and external interventions, and maintaining these dependencies across intervention activities.

Previous studies have examined the role of historical information from multiple perspectives. EgoSchema \citep{mangalam2023egoschema} and HourVideo \citep{chandrasegaran2024hourvideo} focus on video understanding over long time spans, but long-video understanding does not by itself establish how history constrains a task-grounded action. Similarly, long embodied tasks do not necessarily rule out decisions based on current observations. EgoTaskQA \citep{jia2022egotaskqa} establishes task dependencies based on state changes and action conditions; EgoPlan-Bench2 \citep{qiu2026egoplan2} screens the historical requirements in the next-step planning through textual priors and the current frame. FindingDory \citep{yadav2026findingdory} constructs tasks that rely on previous interaction experiences. Yet joint evaluation of the persistent consequences of self-induced anomalies and external interventions, together with paired action selection and causal attribution, remains limited. Existing benchmarks rarely isolate whether an embodied model remembers the causal event that explains why the current task state requires a particular action.

We introduce \textbf{EMBER-Bench}, an \textbf{E}gocentric \textbf{M}emory \textbf{B}enchmark for \textbf{E}mbodied \textbf{R}easoning over cross-event causality in long-horizon household tasks, for which we newly created the task design, video recording, and data annotation. The benchmark evaluates memory utilization through questions at fixed decision points in pre-recorded videos of human behavior, rather than directly testing the robot's closed-loop execution. Memory dependence and answer uniqueness guide construction and review: questions answerable from the goal and current frame alone are revised or excluded, and the available pre-decision history must distinguish the correct answer from the distractors. These memory requirements are reflected in four scenario types. \textbf{L1 (Task Progress Memory)} requires tracking completed steps. \textbf{L2 (Self-induced Anomaly Memory)} concerns recovery obligations arising from errors in the task performer's own actions, requiring models to select recovery actions that address lingering anomalies before normal task execution resumes. \textbf{L3 (External Intervention Memory)} concerns third-party changes to the location, identity, or availability of task-relevant objects, requiring models to update their understanding of object states and adjust current plans. \textbf{L4 (Compound Long-Horizon Memory)} concerns long-horizon tasks that combine self-induced anomalies and external interventions
within the same task episode to evaluate the maintenance and updating of multiple historical constraints across intervening activities.

We construct two complementary question types: \textbf{next-action prediction} and \textbf{causal traceback}. Next-action prediction requires the model to combine the objective, current observation, and relevant history to select one of the candidate actions. For the subset of L2--L4 questions with a uniquely traceable abnormal event, causal traceback provides the reference next action and asks which historical event supports it through a still-valid consequence or task constraint. At matched decision points, we examine whether correct action-cued traceback coincides with correct next-action prediction.

Our contributions are threefold:
\begin{itemize}

    \item We introduce \textbf{EMBER-Bench}, a newly created egocentric benchmark covering task progress, self-induced anomalies, external interventions, and compound long-horizon dependencies, with careful construction and review criteria for memory dependence and answer uniqueness.
    \item We pair next-action prediction with causal traceback at 151 matched decision points. Input ablations localize the key bottleneck: action history alone yields limited gains, whereas explicit causal information yields substantially larger gains.
    \item Across 16 models, we reveal a gap that prior benchmarks do not quantify: correct causal traceback does not necessarily translate into correct next-action prediction. Models can identify the cause given the action, yet often fail to exploit its consequences when predicting what to do next, and simple causal chain-of-thought prompting already yields initial gains.
\end{itemize}
\section{Related Work}
\label{sec:related_work}
\subsection{Long Video Understanding and Egocentric Memory Evaluation}
\label{sec:vlm_memory}
Long-video understanding evaluation focuses on whether the models can locate and integrate information distributed over a relatively long time. EgoSchema \citep{mangalam2023egoschema}, HourVideo \citep{chandrasegaran2024hourvideo}, and LongVideoBench \citep{wu2024longvideobench} provide a basis for long-horizon context evaluation. EgoThink \citep{cheng2024egothink} offers image-level egocentric perception, reasoning and planning evaluations. In terms of historical memory, Ego4D \citep{grauman2022ego4d} adds episodic memory into its core tasks, and QAEgo4D \citep{barmann2022qaego4d} builds memory QA pairs based on Ego4D and examines the utilization of historical information under restricted storage conditions. The EgoMemoria benchmark introduced by MM-Ego \citep{ye2025mmego} further focuses on the perception and memory of visual details in egocentric videos of varying lengths.
A longer time scope brings about the demand for information integration across events and time periods. EgoLife \citep{yang2025egolife}, X-LeBench \citep{zhou2025xlebench}, and EgoMemReason \citep{wang2026egomemreason} have gone beyond the retrieval of a single historical fragment to study the correlation and integration of scattered information. Under continuous input conditions, TeleEgo \citep{yan2025teleego}, EgoSAT \citep{lei2026egosat}, and EOC-Bench \citep{yuan2025eocbench} examine recall and prediction with history. 
Together, these studies examine the retrieval and integration of historical visual information as well as reasoning across events and temporal scales.
\subsection{Embodied Memory and History-dependent Decision-making Evaluation}
\label{sec:embodied_memory}
Embodied-memory evaluation spans \textbf{memory-dependent task execution} \citep{kwon2025memento,yadav2026findingdory,dai2026robomme}, \textbf{history-conditioned reasoning and planning} \citep{jia2022egotaskqa,qiu2026egoplan2}, and \textbf{action-deviation detection and proactive assistance} \citep{peddi2024captaincook4d,kundu2026pwr,gong2026vinci2} (Table~\ref{tab:related-embodied}). However, joint evaluation of how anomalies from different sources continue to constrain subsequent actions after the triggering events leave view, and whether models can trace a given action back to its supporting historical events, remains limited. To address this deficiency, \textbf{EMBER-Bench} covers both types of abnormal sources in the self-designed and collected household tasks: self-induced anomalies and external interventions. Also, \textbf{EMBER-Bench} focuses on persistent action constraints arising from self-induced anomalies and external interventions, and pairs next-action prediction with action-cued causal traceback.
\begin{table*}[t]
\centering
\caption{Comparison with embodied-memory and long-horizon embodied-reasoning benchmarks.}
\label{tab:related-embodied}

\begingroup

\renewcommand{\rwWork}[2]{%
  \textbf{#1}~\citep{#2}%
}

\fontsize{6.2}{7.2}\selectfont

\setlength{\tabcolsep}{1.0pt}
\renewcommand{\arraystretch}{1.00}

\begin{tabularx}{\textwidth}{
@{}
L{0.280\textwidth}
C{0.070\textwidth}
L{0.135\textwidth}
C{0.026\textwidth}
C{0.026\textwidth}
C{0.026\textwidth}
C{0.046\textwidth}
Y
@{}
}
\toprule
\textbf{Benchmark} &
\textbf{Data} &
\textbf{Hist. ctrl.} &
\textbf{L1} &
\textbf{L2} &
\textbf{L3} &
\textbf{L4} &
\textbf{Focus} \\
\midrule

\rwWork{MEMENTO}{kwon2025memento}
& Sim. & Task const. & \N & \N & \N & \N
& Personalized object semantics and user routines. \\

\rwWork{FindingDory}{yadav2026findingdory}
& Sim. & Task const. & \Y & \N & \N & \N
& Conditional, ordered, multi-goal memory control. \\

\rwWork{MIKASA-Robo}{cherepanov2026mikasa}
& Sim.+real & Task const. & \Y & \N & \benchPartial & \N
& Object, spatial, sequence, and capacity memory. \\

\rwWork{RoboMME}{dai2026robomme}
& Sim.+real & Task const. & \Y & \benchPartial & \benchPartial & \benchPartial
& Temporal, spatial, object, and procedural memory. \\

\rwWork{RMBench}{chen2026rmbench}
& Sim.+real & Task const. & \Y & \N & \N & \N
& Manipulation tasks organized by memory complexity. \\

\rwWork{RoboMemArena}{lei2026robomemarena}
& Sim.+real & Task const. & \Y & \N & \benchPartial & \N
& Memory-driven robotic subtasks with keyframes. \\

\rwWork{EgoPlan-Bench}{chen2026egoplan}
& Reuse & Hist. prog. & \Y & \N & \N & \N
& Next-action planning from goal, history, and observation. \\

\rwWork{EgoPlan-Bench2}{qiu2026egoplan2}
& Reuse & Txt./frame filt. & \Y & \N & \N & \N
& Screens text-only and current-frame shortcuts. \\

\rwWork{VidEgoThink}{cheng2024videgothink}
& Reuse & -- & \Y & \benchPartial & \N & \N
& VQA, planning, grounding, and reward modeling. \\

\rwWork{ET-Plan-Bench}{zhang2025etplan}
& Sim. & Temp./preq. & \Y & \benchPartial & \N & \N
& Preconditions, constraints, and replanning. \\

\rwWork{RoboVQA}{sermanet2024robovqa}
& New & Ctxt. plan. & \Y & \benchPartial & \N & \N
& Trajectories, plans, remaining steps, and success. \\

\rwWork{EmbodiedBench}{yang2025embodiedbench}
& Sim. & -- & \benchPartial & \benchPartial & \N & \N
& General embodied decision making. \\

\rwWork{HoloAssist}{wang2023holoassist}
& New & -- & \benchPartial & \benchPartial & \N & \N
& Mistakes and human assistance in collaboration. \\

\rwWork{CaptainCook4D}{peddi2024captaincook4d}
& New & -- & \benchPartial & \benchPartial & \N & \N
& Procedural errors and multistep localization. \\

\rwWork{PREGO}{flaborea2024prego}
& Reuse & Seq. pred. & \benchPartial & \benchPartial & \N & \N
& Online procedural mistake detection. \\

\rwWork{EgoProactive}{kundu2026pwr}
& New & -- & \Y & \Y & \N & \N
& Guidance under deviations and recovery. \\

\rwWork{ProAssist}{zhang2025proassist}
& Reuse & Stream. & \benchPartial & \benchPartial & \N & \N
& Proactive assistance timing and correction. \\

\rwWork{EgoServe / Vinci2}{gong2026vinci2}
& Reuse & Mem. horiz. & \Y & \Y & \N & \N
& Assistance from immediate to cross-session context. \\

\midrule
\rwOwn
& New
& Hist. mask + review
& \Y & \Y & \Y & \Y
& Cross-event state updates and bidirectional causal QA. \\
\bottomrule
\end{tabularx}

\vspace{1mm}
\parbox{\textwidth}{\footnotesize
\textbf{Notes.}
\textbf{Data}: New / Reuse / Sim. / Sim.+real.
\textbf{Hist. ctrl.}: history-control mechanism.
\textbf{L1--L4}: task progress memory, self-induced anomaly memory, external intervention memory, and compound long-horizon memory combining self-induced anomalies and external interventions within a task.
\Y: explicitly evaluated;
\benchPartial: related or partial coverage;
\N: not explicitly targeted.
}
\endgroup
\vspace{-2mm}
\end{table*}

\section{Benchmark Design}
\label{sec:benchmark}
\subsection{Task Design and Data Collection}
\label{sec:task_design}
\textbf{EMBER-Bench} is an original diagnostic benchmark for cross-event memory in egocentric long-horizon household tasks. The tasks for this study were newly created. We also record the corresponding task videos and construct subtask annotations and QA pairs, without splicing or adapting videos, annotations or QA samples from other benchmarks. We adopt a diagnostic assessment approach \citep{johnson2017clevr,patraucean2023perceptiontest} that organizes tests around target competencies, using ``memory necessity for decision-making'' as a guideline throughout the design and review phases.

We target decision points where the goal and current observation are insufficient to select the next action: relevant history must resolve task progress, object states, or action prerequisites. Thus, we construct EMBER-Bench through a detailed pipeline (Figure~\ref{fig:overview}).

We design four types of memory scenarios that these tasks assess, mentioned in the introduction (Figure~\ref{fig:teaser}). Before we capture the video, we need to make sure each task plan of L1--L3 consists of more than 10 atomic actions and more than 40 atomic actions for that of L4. These length thresholds are used only to ensure a non-trivial interaction horizon and are not treated as evidence of memory dependence; the latter is verified separately through the memory-dependence checks described in Section~\ref{sec:quality_control}. The collection of videos covers the kitchen, dining room, living room and bedroom, continuously recording egocentric household tasks. After capturing the videos, we conduct a quality-control pass to ensure adequate visual clarity and verify the interacted objects as well as anomalous or intervention events captured in the recordings.
\subsection{Event and Causal Dependency Annotation}
\label{sec:causal_annotation}
\begin{figure}[t]
    \centering
    \includegraphics[width=1\linewidth]{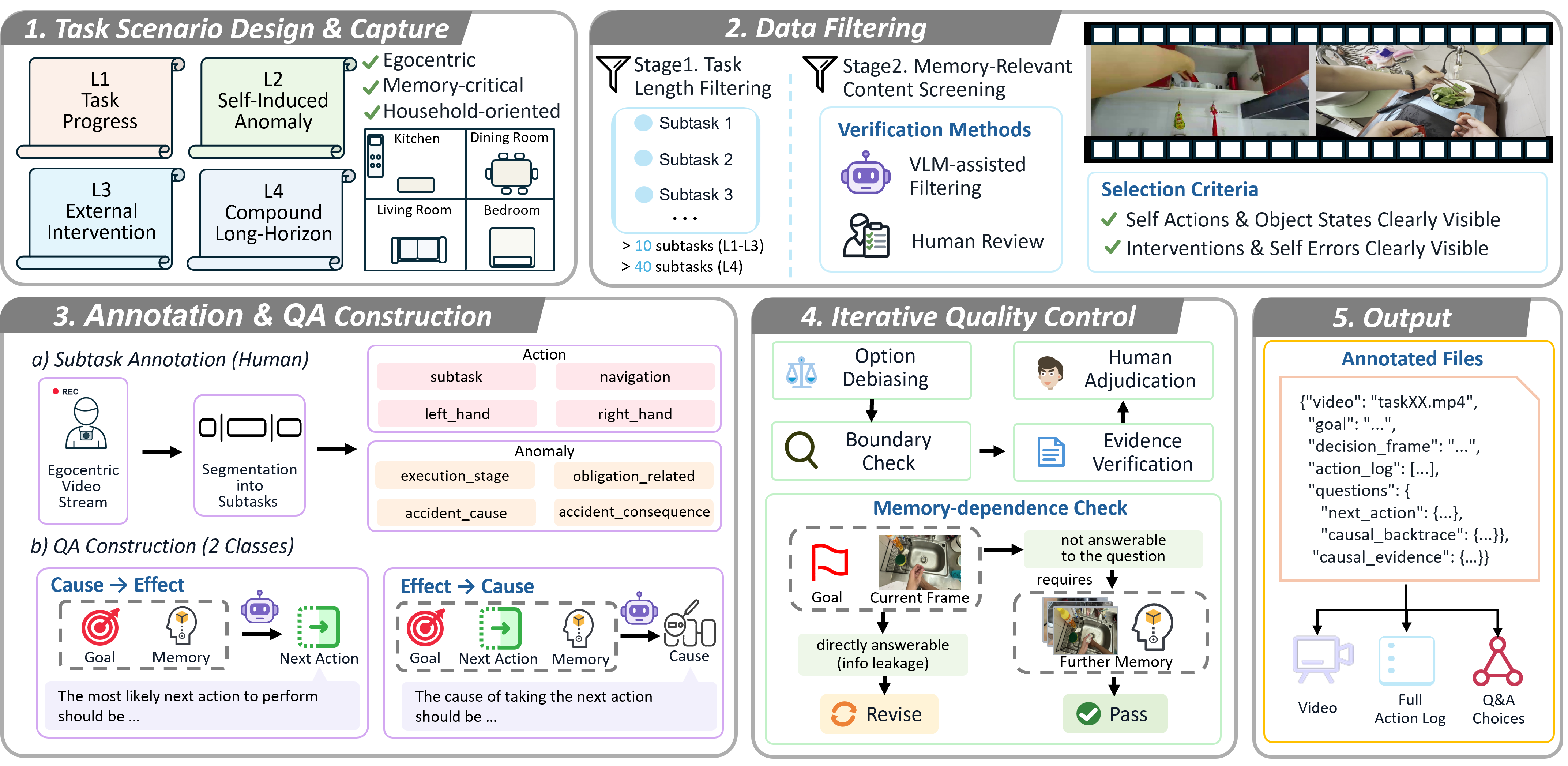}
    \vspace{-10pt} %
    \caption{\textbf{The process of EMBER-Bench construction.} A human-led, model-assisted workflow covers recording, filtering, annotation, and bidirectional QA construction. Iterative quality control checks memory dependence and answer uniqueness, producing traceable benchmark samples.}
    \label{fig:overview}
    \vspace{-10pt} %
\end{figure}

Temporal action labels alone do not specify which past event supports a later recovery action. We therefore augment manual annotations of subtask boundaries, locomotion (\textit{navigation}), and hand interactions (\textit{left\_hand} and \textit{right\_hand}) with structured causal dependencies (Figure~\ref{fig:overview}(3a)). For anomaly-related samples in L2--L4, we record the event phase (\textit{execution\_stage}), the triggering event and its effects (\textit{accident\_cause} and \textit{accident\_consequence}), and the link between a recovery action and the anomaly it addresses (\textit{obligation\_related}). These annotations connect anomalous events, their persistent consequences, and subsequent recovery actions, providing traceable evidence for bidirectional question construction. The cause-and-consequence descriptions also serve as privileged inputs in our information ablation, rather than in the main video-only evaluation. Field definitions, stage-dependent rules, and worked examples of detailed annotations for anomalies are provided in \hyperref[app:action_annotation]{Appendix~\ref*{app:action_annotation}}.
\subsection{Question Design and Input Protocol}
\label{sec:qa_design}

 Based on the event correlation annotation in Section~\ref{sec:causal_annotation}, we transform the continuous impact of historical events on the current task state into two types of complementary causal questions: \textbf{next-action prediction} and \textbf{causal traceback} questions (Figure~\ref{fig:overview}(3b)). By examining the cause-to-effect relationship, we assess whether vision-language models (VLMs) can use historical constraints to decide the next action, whereas by tracing back from the effect to the cause, we examine whether VLMs can trace the core historical basis for a given action. The problems are constructed based on the correlation of \textbf{historical events $\rightarrow$ event consequences $\rightarrow$ subsequent task conditions}. We manually construct the candidate QA pairs and answer choices, have VLMs verify them against the videos and annotations, and finally review them again manually.

First, we construct the \textbf{next-action prediction} multiple-choice questions which allow the tested models to choose one of four options. Each question has a fixed task objective and decision time (frame). The tested models are required to select the next action to be taken at that moment based on the history of previous actions and the current observation. The basis for judgment includes the confirmed status of the incident, its lingering consequences, and the prerequisites for action, which together constrain the choice of the correct answer. When constructing distractor options, we do not simply select actions that have already been completed; rather, we select actions that a model might choose if it misinterprets, ignores, or overreacts to the relevant memory. The \textbf{causal traceback} multiple-choice questions are constructed solely from the subset of next-action prediction questions that have a clear abnormal causal relationship. Some next-action prediction questions rely on the combined history formed by multiple completed steps. Although the next action can be determined, there may not be an explicit reason event suitable for a separate question, so they are retained as next-action prediction questions without being forced to be paired. For eligible L2--L4 prediction questions, we trace the annotated link from the recovery action back to the anomaly it addresses. The associated cause-and-consequence annotations identify the supporting historical event, which becomes the reference answer for the paired causal traceback question (\hyperref[app:action_annotation]{Appendix~\ref*{app:action_annotation}}). In traceback questions, each option is a history clip labeled OPTION A--D in the top-left corner of the video.

The main evaluation provides the task goal, the current decision frame, a pre-decision history video sampled at 5 fps, the question, and four choices. History ends before the reference next action begins. Causal traceback additionally provides that action, with four candidate event clips marked A--D in the history video. Text-input ablations apply only to the 151 paired prediction questions (Section~\ref{sec:memory_quality}). Full prompts and option-rendering details are provided in \hyperref[app:prompts]{Appendix~\ref*{app:prompts}}.
\subsection{Quality Control}
\label{sec:quality_control}
We set the validity of the evidence, answer uniqueness, content clarity, and historical necessity as the pass criteria. First, we need to ensure that the complete video inputs provide only the observed portion (up to the start of the target next action) and do not reveal the future endpoint or subsequent results. Questions answerable from the task goal, current frame, question, and options alone are returned for revision. We conducted three rounds of testing on several mainstream VLMs, providing only the question, choices, and goal, under the condition that no model achieved full correctness across rounds. With the task goal, current frame, question, and choices but no history, accuracy is 26.49\% (shown in \hyperref[app:validation_test]{Appendix~\ref*{app:validation_test}}), compared with 25\% chance accuracy. For each pass question, it should be possible to find the basis that supports the correct option and excludes the other candidates after adding the pre-decision history.

Specific reviews include checks for option debiasing, input boundary check, and memory-dependence check. We control the distribution of correct answer positions, examine the length, wording and detail granularity of the options to identify the cues, confirming that the event boundaries and input text comply with the same time limit as visual input. Through evidence verification, we reference the raw videos and action annotations to guarantee the historical facts supporting the correct answer and to check whether subsequent actions have altered the relevant circumstances. VLM assists in the initial screening, while disputed samples are checked manually based on the original video. The revised questions are re-examined, and samples that still fail to meet the requirements will be excluded.
\subsection{Dataset Statistics and Distribution Analysis}
\label{sec:dataset_statistics}
\begin{figure}
    \centering
    \includegraphics[width=1\linewidth]{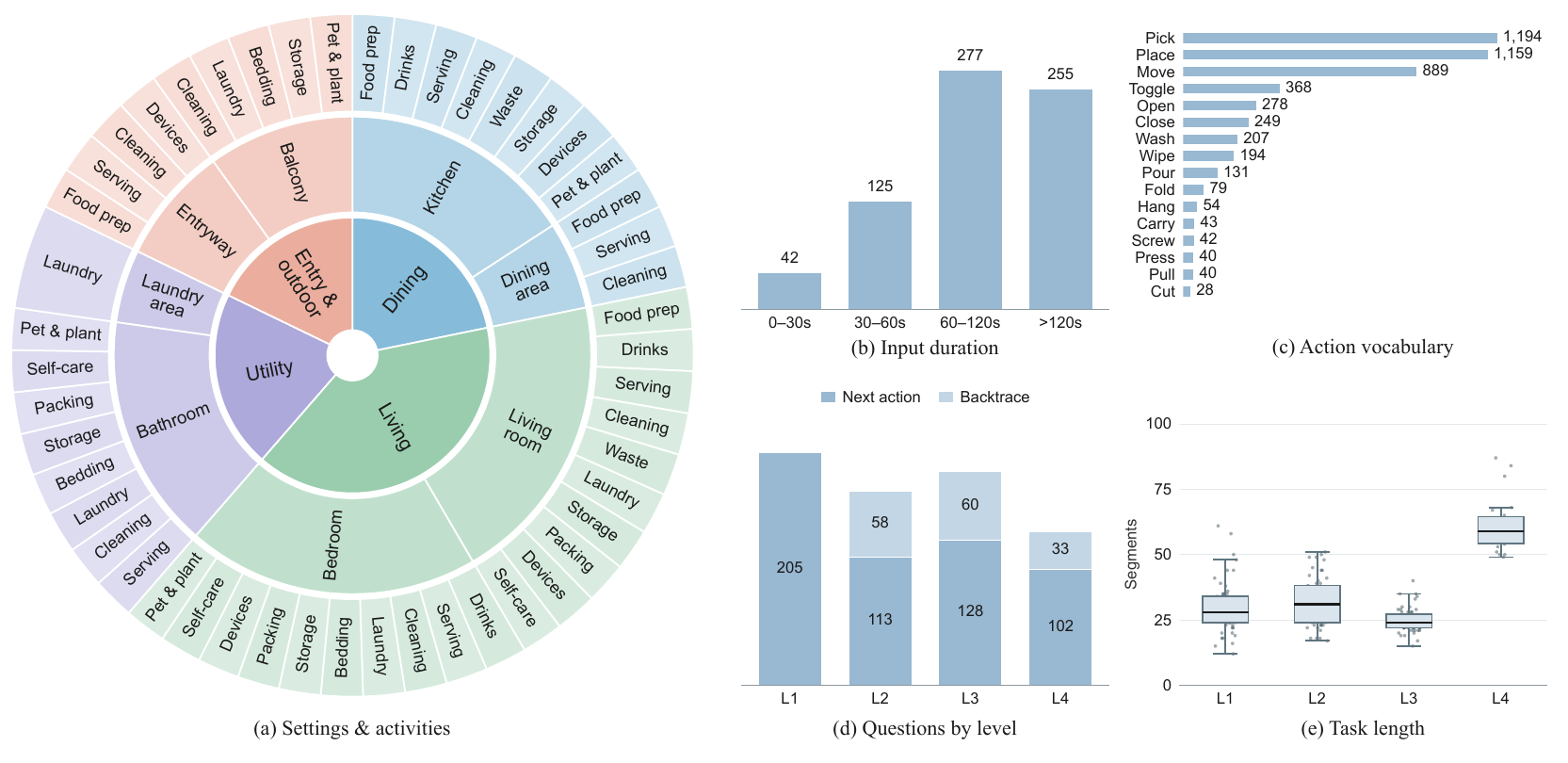}
    \vspace{-10pt}
    \caption{\textbf{Data statistics and distribution of EMBER-Bench.} \textbf{(a)} The hierarchical distribution of scenes and activities; \textbf{(b)} The distribution of video input duration based on all QA pairs; \textbf{(c)} The frequency of the 16 most commonly used types of action words; \textbf{(d)} The number of the two types of problems in L1 to L4; \textbf{(e)} The distribution of task length of L1 to L4 (segments).}
    \vspace{-10pt}
    \label{fig:statistics}
\end{figure}
EMBER-Bench contains a total of 189 self-designed and recorded household tasks, constructing 699 QA pairs, among which 548 are next-action prediction questions and 151 are causal traceback questions. The total duration of the 189 original videos is approximately 9 hours, 13 minutes, and 21 seconds. Our data production team consists of 12 members, whose work covers task design, video collection, action annotation, data filtering, QA construction, and quality verification. The dataset size reflects a selective diagnostic design. Not every long-horizon task or decision point yields a question whose uniquely supported answer requires still-relevant historical evidence beyond the goal and current frame. We therefore prioritize evidence verification, shortcut screening, and iterative revision or exclusion over maximizing task or QA counts (Section~\ref{sec:quality_control}). Causal traceback is further restricted to L2--L4 cases with a uniquely traceable abnormal event (Section~\ref{sec:qa_design}), so not every next-action question produces a paired traceback question.
As shown in Figure~\ref{fig:statistics}(a), the tasks cover living spaces such as the kitchen, dining area, and bedroom, involving household scenarios such as meal preparation, cleaning, organizing, and laundry. L1 to L4 contain 205, 171, 188 and 135 QA pairs, respectively. Next-action prediction questions cover all four scenarios, while causal traceback questions cover only L2 to L4 (numbers are shown in Figure~\ref{fig:statistics}(d)). Figure~\ref{fig:statistics}(b) shows the video input duration. Among them, there are 532 samples exceeding 60 seconds, accounting for 76.1\% in total. %
\section{Experiments}
\label{sec:experiments}
We evaluate 16 multimodal large language models (MLLMs; \hyperref[app:baseline_models]{Appendix~\ref*{app:baseline_models}}) on four questions: dependency difficulty, the information that changes decisions, the cause--action gap, and whether explicit reasoning closes it. For video inputs, we uniformly sample the provided clips at a target rate of 5 fps. We use a common prompt per question type (\hyperref[app:prompts]{Appendix~\ref*{app:prompts}}). The evaluation protocol is detailed in \hyperref[app:exp_protocol]{Appendix~\ref*{app:exp_protocol}}. The main setting, denoted by $V$, provides only the task goal, the pre-decision history video, and the current frame.
\subsection{How Well Do Models Handle Each Memory Dependency?}
\label{sec:overall_performance}
Table~\ref{tab:main} reports accuracy by dependency level. Each question has four options (25\% chance accuracy). Two human evaluators reach 98.3\% overall (97.8\% on $P$, 100\% on $C$; \hyperref[app:human]{Appendix~\ref*{app:human}}); the best model, Gemini 3.8 Flash, achieves 61.2\% ($P$ 58.4\%, $C$ 71.5\%). Averaged over models, $P$ accuracy is 47.1\%, 35.3\%, 36.6\%, and 31.0\% for L1--L4, respectively; 12 of the 16 models perform worst on L4. Fifteen of the 16 models score higher on $C$ than on $P$; the exception, Qwen3-VL 8B, is near chance on $C$ (25.8\%). Across models, $P$ and $C$ accuracy correlate at $r = 0.85$, and $P$ accuracies at different levels correlate at 0.76--0.94. The higher $C$ accuracy may partly reflect the action cue that $C$ provides.

\begin{figure*}[t]
\centering
\includegraphics[width=0.9\textwidth]{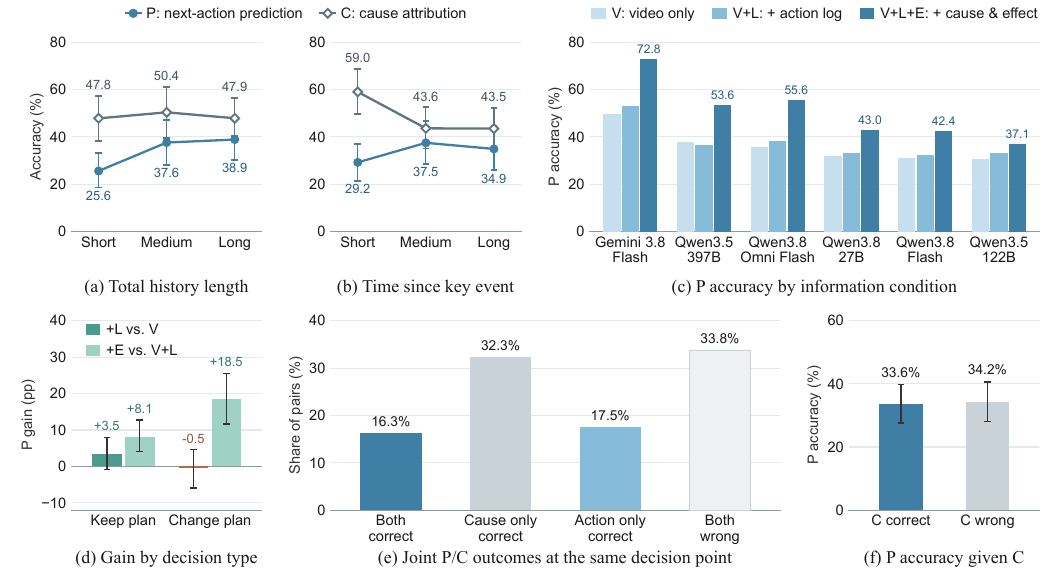}
 \vspace{-10pt}
\caption{Temporal span, information ablation, and paired prediction--traceback outcomes. (a--b) $P$ and $C$ accuracy across tertiles of total history duration and event-to-decision interval. (c--d) $P$ accuracy of six models under $V$, $V{+}L$, and $V{+}L{+}E$ inputs, and gains by decision type. (e--f) Joint $P$/$C$ outcomes and $P$ accuracy conditioned on $C$ correctness. Details: \hyperref[app:stats]{Appendix~\ref*{app:stats}}.}
 \vspace{-10pt}
\label{fig:exp_analysis}

\end{figure*}

For the 151 matched pairs (99 tasks), we split total history duration and the event-to-decision interval (from the end of the key event) separately into tertiles (\hyperref[app:temporal]{Appendix~\ref*{app:temporal}}). $P$ rises from 25.6\% to 38.9\% from the shortest to the longest history group, while $C$ stays at 47.8--50.4\%. As the interval grows, $C$ falls from 59.0\% to 43.5\%, whereas $P$ shows no corresponding decline (Figure~\ref{fig:exp_analysis}(a--b)). In a regression controlling for dependency level and both time variables, each doubling of total history is associated with about $+11.7$ points in $P$, and each doubling of the interval (plus 1\,s) with about $-3.7$ points in $C$; the other two associations are not significant (\hyperref[app:temporal]{Appendix~\ref*{app:temporal}}). More distant events are therefore harder to trace, but do not necessarily correspond to lower prediction accuracy. The positive association with total duration may reflect task structure and should not be interpreted as evidence that extending the history improves decisions.

\begin{table*}[t]
\centering
\caption{Accuracy (\%) under the video-history setting $V$. $P$ denotes next-action prediction and $C$ causal traceback. Models are grouped into closed-source and open-source models and sorted by overall accuracy within each group. The Human row averages two evaluators (\hyperref[app:human]{Appendix~\ref*{app:human}}) and is excluded from ranking. The best model result in each column is shaded and bold; the second best is underlined. Unanswered items after retries are scored incorrect. Random-choice accuracy is 25\%.}
\label{tab:main}
\includegraphics[width=\textwidth]{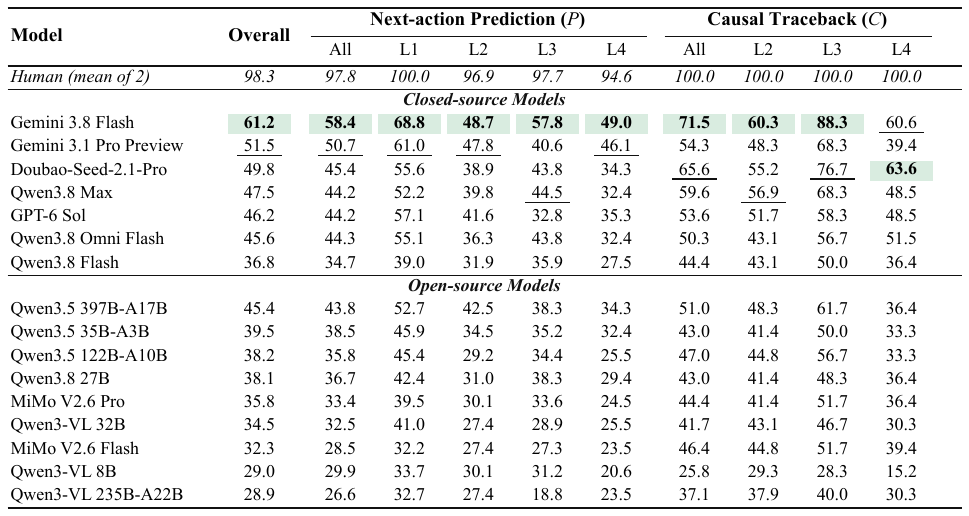}
\end{table*}

\subsection{Which Historical Information Changes Decisions?}
\label{sec:memory_quality}

We progressively add two kinds of historical information to 151 prediction questions with explicit causal chains: action records and event consequences. The $V{+}L$ setting adds action descriptions, timestamps, and hand or navigation logs of subtasks completed before the decision point; the $V{+}L{+}E$ setting further adds cause and consequence annotations for completed events. Here, $E$ is privileged information used for diagnosis; performance under this setting does not measure independent causal inference. We run this ablation on six models spanning the capability range of the leaderboard. Averaged over the six models, action logs add only 1.6 points, whereas cause-and-consequence annotations add a further 13.0 points (Figure~\ref{fig:exp_analysis}(c)). The annotation gain is 18.5 points for decisions that change the original plan (71 questions) and 8.1 points for those that keep it (80 questions), a difference of about 10.4 points (Figure~\ref{fig:exp_analysis}(d); \hyperref[app:ablation]{Appendix~\ref*{app:ablation}}). For anomaly-dependent decisions, explicitly specifying how events constrain subsequent actions is thus more helpful than recovering the sequence of completed steps alone.

Annotation gains vary across models: 19.9 points for Gemini 3.8 Flash, 17.2 for Qwen3.5 397B-A17B and Qwen3.8 Omni Flash, 9.9 for Qwen3.8 27B and Qwen3.8 Flash, and only 4.0 (not significant) for Qwen3.5 122B-A10B (\hyperref[app:ablation]{Appendix~\ref*{app:ablation}}). Given the same annotations, models differ in the degree to which they convert them into the correct action.

\subsection{Does Identifying a Cause Coincide with the Right Action?}
\label{sec:joint_analysis}

We pair video-only $P$ and $C$ responses from 16 models at the 151 matched decision points. Both answers are correct in 16.3\% of the 2,416 paired responses, only $C$ in 32.3\%, only $P$ in 17.5\%, and neither in 33.8\% (Figure~\ref{fig:exp_analysis}(e)). This joint rate is close to the 16.5\% expected under independence between $P$ and $C$ correctness; its low level is driven mainly by the low marginal $P$ accuracy. $P$ accuracy is 33.6\% when $C$ is correct and 34.2\% when it is not, a difference of about $-0.6$ points; controlling for model gives about $-2.7$, and additionally controlling for question pair, which absorbs item difficulty, gives about $-3.5$ (\hyperref[app:joint]{Appendix~\ref*{app:joint}}; Figure~\ref{fig:exp_analysis}(f)). None of these differences is significant, and the most adjusted estimate rules out a positive association larger than about 0.6 points. Correctly identifying the relevant event in a traceback question is thus not accompanied by a detectable gain in choosing the correct action at the same decision point. One possible explanation, consistent with this result and with the annotation gains in Section~\ref{sec:memory_quality}, is that models do not spontaneously exploit an event's persisting effects when selecting the next action. This interpretation requires a caveat: because $C$ provides the action as a cue, models may solve it by matching the action to a candidate clip rather than by retrieving its cause; the results therefore establish a behavioral mismatch rather than a failure of internal causal retrieval.

\subsection{Can Explicit Causal Reasoning Close the Gap?}
\label{sec:causal_reasoning}

To close this gap, we design a simple causal chain-of-thought (CoT) \citep{wei2022chain,kojima2022large} prompt that asks models to first assess whether historical events changed current conditions and whether their effects remain relevant, then select an action, without additional annotations or traceback answers (prompt in \hyperref[app:prompts]{Appendix~\ref*{app:prompts}}). Because CoT substantially increases inference cost, we evaluate three models: the strongest model, Gemini 3.8 Flash, and two Qwen models of different scales (Qwen3.8 Flash and Qwen3.8 27B). The direct baseline of each model is re-measured in the same run as its CoT condition. Gemini 3.8 Flash improves from 47.0\% to 55.0\% (21 answers corrected, 9 newly wrong; a significant gain of about 7.9 points) and gains 13.0 points within the correct-$C$ group (Table~\ref{tab:exp_cot}; examples in \hyperref[app:cot]{Appendix~\ref*{app:cot}}). For this model, the difference in $P$ accuracy between correct-$C$ and incorrect-$C$ pairs rises from about $-12.3$ points under direct answering to about $+5.3$ points under CoT, a significant change of about 17.6 points (\hyperref[app:cot]{Appendix~\ref*{app:cot}}). Explicitly reasoning about event consequences can therefore alleviate some cases of correct traceback paired with incorrect decisions. Qwen3.8 Flash changes from 31.1\% to 34.4\% and Qwen3.8 27B from 31.8\% to 27.8\%; neither change is significant. The overall drop of Qwen3.8 27B comes from questions outside the correct-$C$ group; within that group, corrected and newly wrong answers cancel (net 0.0 in Table~\ref{tab:exp_cot}). This pattern is consistent with a capability-dependent effect: stronger models may convert intermediate judgments about event effects into correct actions more reliably, whereas weaker models may amplify flawed judgments into confidently wrong ones. With only three models, however, this interpretation remains speculative and requires further verification; a systematic analysis of intermediate reasoning quality is left to future work.

\begin{table}[t]
\centering
\setlength{\abovecaptionskip}{5pt}\caption{Decision gains from causal CoT guidance. The final column is the net gain within the correct-$C$ group (fixed by main-evaluation results), $100\times(\text{corrected}-\text{newly wrong})/(\text{group size})$, comparing CoT against direct answering on the same 151 questions; it is not the overall direct-to-CoT change.}
\label{tab:exp_cot}
\setlength{\tabcolsep}{9pt}\small\begin{tabular}{lrrr}
\toprule
Model & Direct (\%) & Causal CoT (\%) & Net gain, correct-$C$ group (pp) \\
\midrule
Gemini 3.8 Flash & 47.0 & 55.0 & +13.0 \\
Qwen3.8 Flash & 31.1 & 34.4 & +4.5 \\
Qwen3.8 27B & 31.8 & 27.8 & 0.0 \\
\bottomrule
\end{tabular}
\vspace{-\baselineskip}
\end{table}

\section{Conclusion}
\label{sec:conclusion}
Long-horizon embodied tasks demand more than processing longer histories: agents must identify causally relevant events and understand how their persistent consequences constrain current actions. To address the limited evaluation of this capability, we introduce EMBER-Bench, an egocentric benchmark spanning task progress, self-induced anomalies, external interventions, and compound long-horizon dependencies. Paired next-action prediction and action-cued causal traceback examine history-grounded action selection and its supporting historical events. Paired analyses show that correct traceback does not reliably coincide with correct next-action prediction; input ablations show that causal annotations, not action logs, improve decisions, and a simple causal chain of thought yields initial gains. The benchmark is limited to offline multiple-choice evaluation on pre-recorded household tasks. Future work should extend evaluation to closed-loop interaction and develop memory systems that track persistent consequences and update action constraints, supporting recovery and replanning in lifelong physical agents.
\subsection*{AI Use Statement}
Generative AI tools were used during paper preparation for language polishing, typo and grammar correction, and LaTeX formatting.

During data construction, task design, video recording, subtask annotation, and candidate QA construction were carried out by the research team. Vision-language models assisted only in verifying candidate QA pairs and in the initial quality-control screening, with reference to the raw videos and human annotations. All candidate QA pairs were subsequently reviewed by human evaluators, and disputed samples were re-examined against the original videos; samples that still failed the criteria for evidence validity, answer uniqueness, and memory dependence after revision were excluded. Model outputs were never the sole basis for determining a reference answer or retaining a sample.

The authors reviewed, revised, and verified all AI-assisted content and bear full responsibility for the study design, data quality, experimental results, and conclusions of the paper.

\subsection*{Ethics Statement}
EMBER-Bench is intended to support research on history-dependent reasoning in embodied agents. Its household tasks, egocentric recordings, and annotations were newly created by the research team. The L3 and L4 scenarios involve filming human participants, with faces visible in some of the original recordings. Video participants agreed to participate in the study. To protect participants' privacy, we blur identifiable faces in the videos. Model evaluations are conducted offline on prerecorded videos rather than through autonomous robot operation. Benchmark performance should not be interpreted as evidence of safe or reliable deployment in real homes.
\subsection*{Reproducibility Statement}
To support reproducibility, Section~\ref{sec:benchmark} describes task design, video collection, annotation, question construction, and quality control. \hyperref[app:action_annotation]{Appendix~\ref*{app:action_annotation}} details the annotation schema and paired-question construction. \hyperref[app:prompts]{Appendix~\ref*{app:prompts}} provides the evaluation prompts, input specifications for each condition, and answer-parsing rules, while \hyperref[app:validation_test]{Appendix~\ref*{app:validation_test}} reports the current-frame-only control. \hyperref[app:baseline_models]{Appendix~\ref*{app:baseline_models}} lists the evaluated models, and \hyperref[app:experiments]{Appendix~\ref*{app:experiments}} describes the evaluation protocol, analysis subsets, statistical procedures, and human evaluation. These sections document how benchmark questions are constructed, presented to models, and scored, together with the procedures used for the reported analyses. The project is available at \href{https://zhaoalexgoat.github.io/EMBER-Bench/}
{\textcolor{urlteal}{\nolinkurl{https://zhaoalexgoat.github.io/EMBER-Bench/}}}.

\raggedbottom
\bibliography{references_embodied}

\begin{thebibliography}{47}
\providecommand{\natexlab}[1]{#1}
\providecommand{\url}[1]{\texttt{#1}}
\expandafter\ifx\csname urlstyle\endcsname\relax
  \providecommand{\doi}[1]{doi: #1}\else
  \providecommand{\doi}{doi: \begingroup \urlstyle{rm}\Url}\fi

\bibitem[{Alibaba Cloud}(2026{\natexlab{a}})]{alibaba2026qwen38flash}
{Alibaba Cloud}.
\newblock {Qwen3.8-Flash}.
\newblock Model Studio documentation, 2026{\natexlab{a}}.
\newblock URL
  \url{https://www.alibabacloud.com/help/en/model-studio/qwen3-8-flash}.
\newblock Accessed September 25, 2026.

\bibitem[{Alibaba Cloud}(2026{\natexlab{b}})]{alibaba2026qwen38omniflash}
{Alibaba Cloud}.
\newblock {Qwen3.8-Omni-Flash}.
\newblock Model Studio documentation, 2026{\natexlab{b}}.
\newblock URL
  \url{https://www.alibabacloud.com/help/en/model-studio/qwen3-8-omni-flash}.
\newblock Accessed September 25, 2026.

\bibitem[Bai et~al.(2025)Bai, Cai, Chen, Chen, Chen, Cheng, Deng, Ding, Gao,
  Ge, Ge, Guo, Huang, Huang, Huang, Hui, Jiang, Li, Li, Li, Li, Lin, Lin, Liu,
  Liu, Liu, Liu, Liu, Liu, Lu, Luo, Lv, Men, Meng, Ren, Ren, Song, Sun, Tang,
  Tu, Wan, Wang, Wang, Wang, Wang, Xie, Xu, Xu, Xu, Yang, Yang, Yang, Yang, Yu,
  Zhang, Zhang, Zhang, Zheng, Zhong, Zhou, Zhou, Zhou, Zhu, and
  Zhu]{bai2025qwen3vl}
Shuai Bai, Yuxuan Cai, Ruizhe Chen, Keqin Chen, Xionghui Chen, Zesen Cheng,
  Lianghao Deng, Wei Ding, Chang Gao, Chunjiang Ge, Wenbin Ge, Zhifang Guo,
  Qidong Huang, Jie Huang, Fei Huang, Binyuan Hui, Shutong Jiang, Zhaohai Li,
  Mingsheng Li, Mei Li, Kaixin Li, Zicheng Lin, Junyang Lin, Xuejing Liu,
  Jiawei Liu, Chenglong Liu, Yang Liu, Dayiheng Liu, Shixuan Liu, Dunjie Lu,
  Ruilin Luo, Chenxu Lv, Rui Men, Lingchen Meng, Xuancheng Ren, Xingzhang Ren,
  Sibo Song, Yuchong Sun, Jun Tang, Jianhong Tu, Jianqiang Wan, Peng Wang,
  Pengfei Wang, Qiuyue Wang, Yuxuan Wang, Tianbao Xie, Yiheng Xu, Haiyang Xu,
  Jin Xu, Zhibo Yang, Mingkun Yang, Jianxin Yang, An~Yang, Bowen Yu, Fei Zhang,
  Hang Zhang, Xi~Zhang, Bo~Zheng, Humen Zhong, Jingren Zhou, Fan Zhou, Jing
  Zhou, Yuanzhi Zhu, and Ke~Zhu.
\newblock {Qwen3-VL Technical Report}.
\newblock \emph{arXiv preprint arXiv:2511.21631}, 2025.
\newblock URL \url{https://arxiv.org/abs/2511.21631}.

\bibitem[B{\"a}rmann \& Waibel(2022)B{\"a}rmann and Waibel]{barmann2022qaego4d}
Leonard B{\"a}rmann and Alex Waibel.
\newblock {Where Did I Leave My Keys? -- Episodic-Memory-Based Question
  Answering on Egocentric Videos}.
\newblock In \emph{Proceedings of the IEEE/CVF Conference on Computer Vision
  and Pattern Recognition Workshops}, pp.\  1560--1568, 2022.
\newblock URL
  \url{https://openaccess.thecvf.com/content/CVPR2022W/Ego4D-EPIC/html/Barmann_Where_Did_I_Leave_My_Keys_-_Episodic-Memory-Based_Question_Answering_CVPRW_2022_paper.html}.

\bibitem[{ByteDance Seed Team}(2026)]{bytedance2026seed21}
{ByteDance Seed Team}.
\newblock {Seed2.1 Officially Released: Advancing AI Productivity}.
\newblock Official model release, June 2026.
\newblock URL
  \url{https://seed.bytedance.com/en/blog/seed2-1-officially-released-advancing-ai-productivity}.
\newblock Accessed September 25, 2026.

\bibitem[Chandrasegaran et~al.(2024)Chandrasegaran, Gupta, Hadzic, Kota, He,
  Eyzaguirre, Durante, Li, Wu, and Fei-Fei]{chandrasegaran2024hourvideo}
Keshigeyan Chandrasegaran, Agrim Gupta, Lea~M. Hadzic, Taran Kota, Jimming He,
  Cristobal Eyzaguirre, Zane Durante, Manling Li, Jiajun Wu, and Li~Fei-Fei.
\newblock {HourVideo: 1-Hour Video-Language Understanding}.
\newblock In \emph{Advances in Neural Information Processing Systems},
  volume~37, pp.\  53168--53197, 2024.
\newblock \doi{10.52202/079017-1684}.
\newblock URL
  \url{https://proceedings.neurips.cc/paper_files/paper/2024/hash/5f2809607f692d79a01c05c43d702883-Abstract-Datasets_and_Benchmarks_Track.html}.

\bibitem[Chen et~al.(2026{\natexlab{a}})Chen, Wang, Li, Qin, Shi, Li, Hu,
  Zhang, Wang, Chen, Wang, Wang, Yang, Xu, Wu, Mu, Yang, Dong, and
  Luo]{chen2026rmbench}
Tianxing Chen, Yuran Wang, Mingleyang Li, Yan Qin, Hao Shi, Zixuan Li, Yifan
  Hu, Yingsheng Zhang, Kaixuan Wang, Yue Chen, Hongcheng Wang, Junjie Wang,
  Tianhang Yang, Renjing Xu, Ruihai Wu, Yao Mu, Yaodong Yang, Hao Dong, and
  Ping Luo.
\newblock {RMBench: Memory-Dependent Robotic Manipulation Benchmark with
  Insights into Policy Design}.
\newblock \emph{arXiv preprint arXiv:2603.01229}, 2026{\natexlab{a}}.
\newblock URL \url{https://arxiv.org/abs/2603.01229}.

\bibitem[Chen et~al.(2026{\natexlab{b}})Chen, Ge, Ge, Ding, Li, Wang, Xu, Shan,
  and Liu]{chen2026egoplan}
Yi~Chen, Yuying Ge, Yixiao Ge, Mingyu Ding, Bohao Li, Rui Wang, Ruifeng Xu,
  Ying Shan, and Xihui Liu.
\newblock {EgoPlan-Bench: Benchmarking Multimodal Large Language Models for
  Human-Level Planning}.
\newblock \emph{International Journal of Computer Vision}, 134,
  2026{\natexlab{b}}.
\newblock \doi{10.1007/s11263-025-02676-0}.
\newblock URL
  \url{https://link.springer.com/article/10.1007/s11263-025-02676-0}.
\newblock Article 118.

\bibitem[Cheng et~al.(2024{\natexlab{a}})Cheng, Fang, Yu, Zhou, Li, Tian, Li,
  Han, and Liu]{cheng2024videgothink}
Sijie Cheng, Kechen Fang, Yangyang Yu, Sicheng Zhou, Bohao Li, Ye~Tian,
  Tingguang Li, Lei Han, and Yang Liu.
\newblock {VidEgoThink: Assessing Egocentric Video Understanding Capabilities
  for Embodied AI}.
\newblock \emph{arXiv preprint arXiv:2410.11623}, 2024{\natexlab{a}}.
\newblock URL \url{https://arxiv.org/abs/2410.11623}.

\bibitem[Cheng et~al.(2024{\natexlab{b}})Cheng, Guo, Wu, Fang, Li, Liu, and
  Liu]{cheng2024egothink}
Sijie Cheng, Zhicheng Guo, Jingwen Wu, Kechen Fang, Peng Li, Huaping Liu, and
  Yang Liu.
\newblock {EgoThink: Evaluating First-Person Perspective Thinking Capability of
  Vision-Language Models}.
\newblock In \emph{Proceedings of the IEEE/CVF Conference on Computer Vision
  and Pattern Recognition}, pp.\  14291--14302, 2024{\natexlab{b}}.
\newblock URL
  \url{https://openaccess.thecvf.com/content/CVPR2024/html/Cheng_EgoThink_Evaluating_First-Person_Perspective_Thinking_Capability_of_Vision-Language_Models_CVPR_2024_paper.html}.

\bibitem[Cherepanov et~al.(2026)Cherepanov, Kachaev, Kovalev, and
  Panov]{cherepanov2026mikasa}
Egor Cherepanov, Nikita Kachaev, Alexey Kovalev, and Aleksandr Panov.
\newblock {Memory, Benchmark \& Robots: A Benchmark for Solving Complex Tasks
  with Reinforcement Learning}.
\newblock In \emph{International Conference on Learning Representations}, 2026.
\newblock URL
  \url{https://proceedings.iclr.cc/paper_files/paper/2026/hash/1088fffae5f2aa8c8324e4f45f62248d-Abstract-Conference.html}.

\bibitem[Dai et~al.(2026)Dai, Fu, Lee, Liu, Zhang, Yang, Finn, Fazeli, and
  Chai]{dai2026robomme}
Yinpei Dai, Hongze Fu, Jayjun Lee, Yuejiang Liu, Haoran Zhang, Jianing Yang,
  Chelsea Finn, Nima Fazeli, and Joyce Chai.
\newblock {RoboMME: Benchmarking and Understanding Memory for Robotic
  Generalist Policies}.
\newblock In \emph{International Conference on Machine Learning}, 2026.
\newblock URL \url{https://arxiv.org/abs/2603.04639}.

\bibitem[Flaborea et~al.(2024)Flaborea, di~Melendugno, Plini, Scofano,
  De~Matteis, Furnari, Farinella, and Galasso]{flaborea2024prego}
Alessandro Flaborea, Guido Maria~D'Amely di~Melendugno, Leonardo Plini, Luca
  Scofano, Edoardo De~Matteis, Antonino Furnari, Giovanni~Maria Farinella, and
  Fabio Galasso.
\newblock {PREGO: Online Mistake Detection in PRocedural EGOcentric Videos}.
\newblock In \emph{Proceedings of the IEEE/CVF Conference on Computer Vision
  and Pattern Recognition}, pp.\  18483--18492, 2024.
\newblock \doi{10.1109/CVPR52733.2024.01749}.
\newblock URL
  \url{https://openaccess.thecvf.com/content/CVPR2024/html/Flaborea_PREGO_Online_Mistake_Detection_in_PRocedural_EGOcentric_Videos_CVPR_2024_paper.html}.

\bibitem[Gong et~al.(2026)Gong, Yan, Kang, Zheng, Ruan, Lu, Zhang, Sato, and
  Huang]{gong2026vinci2}
Sitong Gong, Tianyu Yan, Caixin Kang, Bo~Zheng, Xiang Ruan, Huchuan Lu, Kaipeng
  Zhang, Yoichi Sato, and Yifei Huang.
\newblock {Vinci2: Providing Proactive Assistance in Continuous Egocentric
  Videos}.
\newblock In \emph{European Conference on Computer Vision}, 2026.
\newblock URL \url{https://arxiv.org/abs/2607.11523}.

\bibitem[{Google}(2026{\natexlab{a}})]{google2026gemini31pro}
{Google}.
\newblock {Gemini 3.1 Pro Preview}.
\newblock Gemini API documentation, 2026{\natexlab{a}}.
\newblock URL
  \url{https://ai.google.dev/gemini-api/docs/models/gemini-3.1-pro-preview}.
\newblock Accessed September 25, 2026.

\bibitem[{Google}(2026{\natexlab{b}})]{google2026gemini38flash}
{Google}.
\newblock {Gemini 3.8 Flash}.
\newblock Gemini API documentation, 2026{\natexlab{b}}.
\newblock URL
  \url{https://ai.google.dev/gemini-api/docs/models/gemini-3.8-flash}.
\newblock Accessed September 25, 2026.

\bibitem[Grauman et~al.(2022)Grauman, Westbury, Byrne, Chavis, Furnari,
  Girdhar, Hamburger, Jiang, Liu, Liu, Martin, Nagarajan, Radosavovic,
  Ramakrishnan, Ryan, Sharma, Wray, Xu, Xu, Zhao, Bansal, Batra, Cartillier,
  Crane, Do, Doulaty, Erapalli, Feichtenhofer, Fragomeni, Fu, Gebreselasie,
  Gonzalez, Hillis, Huang, Huang, Jia, Khoo, Kolar, Kottur, Kumar, Landini, Li,
  Li, Li, Mangalam, Modhugu, Munro, Murrell, Nishiyasu, Price, Puentes,
  Ramazanova, Sari, Somasundaram, Southerland, Sugano, Tao, Vo, Wang, Wu, Yagi,
  Zhao, Zhu, Arbelaez, Crandall, Damen, Farinella, Fuegen, Ghanem, Ithapu,
  Jawahar, Joo, Kitani, Li, Newcombe, Oliva, Park, Rehg, Sato, Shi, Shou,
  Torralba, Torresani, Yan, and Malik]{grauman2022ego4d}
Kristen Grauman, Andrew Westbury, Eugene Byrne, Zachary Chavis, Antonino
  Furnari, Rohit Girdhar, Jackson Hamburger, Hao Jiang, Miao Liu, Xingyu Liu,
  Miguel Martin, Tushar Nagarajan, Ilija Radosavovic, Santhosh~Kumar
  Ramakrishnan, Fiona Ryan, Jayant Sharma, Michael Wray, Mengmeng Xu,
  Eric~Zhongcong Xu, Chen Zhao, Siddhant Bansal, Dhruv Batra, Vincent
  Cartillier, Sean Crane, Tien Do, Morrie Doulaty, Akshay Erapalli, Christoph
  Feichtenhofer, Adriano Fragomeni, Qichen Fu, Abrham Gebreselasie, Cristina
  Gonzalez, James Hillis, Xuhua Huang, Yifei Huang, Wenqi Jia, Weslie Khoo,
  Jachym Kolar, Satwik Kottur, Anurag Kumar, Federico Landini, Chao Li, Yanghao
  Li, Zhenqiang Li, Karttikeya Mangalam, Raghava Modhugu, Jonathan Munro,
  Tullie Murrell, Takumi Nishiyasu, Will Price, Paola~Ruiz Puentes, Merey
  Ramazanova, Leda Sari, Kiran Somasundaram, Audrey Southerland, Yusuke Sugano,
  Ruijie Tao, Minh Vo, Yuchen Wang, Xindi Wu, Takuma Yagi, Ziwei Zhao, Yunyi
  Zhu, Pablo Arbelaez, David Crandall, Dima Damen, Giovanni~Maria Farinella,
  Christian Fuegen, Bernard Ghanem, Vamsi~Krishna Ithapu, C.~V. Jawahar,
  Hanbyul Joo, Kris Kitani, Haizhou Li, Richard Newcombe, Aude Oliva, Hyun~Soo
  Park, James~M. Rehg, Yoichi Sato, Jianbo Shi, Mike~Zheng Shou, Antonio
  Torralba, Lorenzo Torresani, Mingfei Yan, and Jitendra Malik.
\newblock {Ego4D: Around the World in 3,000 Hours of Egocentric Video}.
\newblock In \emph{Proceedings of the IEEE/CVF Conference on Computer Vision
  and Pattern Recognition}, pp.\  18995--19012, 2022.
\newblock URL
  \url{https://openaccess.thecvf.com/content/CVPR2022/html/Grauman_Ego4D_Around_the_World_in_3000_Hours_of_Egocentric_Video_CVPR_2022_paper.html}.

\bibitem[Jia et~al.(2022)Jia, Lei, Zhu, and Huang]{jia2022egotaskqa}
Baoxiong Jia, Ting Lei, Song-Chun Zhu, and Siyuan Huang.
\newblock {EgoTaskQA: Understanding Human Tasks in Egocentric Videos}.
\newblock In \emph{Advances in Neural Information Processing Systems},
  volume~35, pp.\  3343--3360, 2022.
\newblock \doi{10.52202/068431-0242}.
\newblock URL
  \url{https://papers.nips.cc/paper_files/paper/2022/hash/161c94a58ca25bafcaf47893e8233deb-Abstract-Datasets_and_Benchmarks.html}.

\bibitem[Johnson et~al.(2017)Johnson, Hariharan, van~der Maaten, Fei-Fei,
  Zitnick, and Girshick]{johnson2017clevr}
Justin Johnson, Bharath Hariharan, Laurens van~der Maaten, Li~Fei-Fei,
  C.~Lawrence Zitnick, and Ross Girshick.
\newblock {CLEVR: A Diagnostic Dataset for Compositional Language and
  Elementary Visual Reasoning}.
\newblock In \emph{Proceedings of the IEEE Conference on Computer Vision and
  Pattern Recognition}, pp.\  2901--2910, 2017.
\newblock URL
  \url{https://openaccess.thecvf.com/content_cvpr_2017/html/Johnson_CLEVR_A_Diagnostic_CVPR_2017_paper.html}.

\bibitem[Kojima et~al.(2022)Kojima, Gu, Reid, Matsuo, and
  Iwasawa]{kojima2022large}
Takeshi Kojima, Shixiang~(Shane) Gu, Machel Reid, Yutaka Matsuo, and Yusuke
  Iwasawa.
\newblock Large language models are {Zero-Shot} reasoners.
\newblock In \emph{Advances in Neural Information Processing Systems},
  volume~35, pp.\  22199--22213, 2022.
\newblock \doi{10.52202/068431-1613}.
\newblock URL
  \url{https://proceedings.neurips.cc/paper/2022/hash/8bb0d291acd4acf06ef112099c16f326-Abstract-Conference.html}.

\bibitem[Kundu et~al.(2026)Kundu, Shrivastava, Arap, Wang, Zhu, Fettes, Tiwari,
  Suresh, Moutakanni, Castillejo~Munoz, Bolourchi, Fung, Donmez, Damavandi,
  Kumar, and Moon]{kundu2026pwr}
Kaustav Kundu, Ritvik Shrivastava, Maxim Arap, Nanshu Wang, Xianhui Zhu,
  Quintin Fettes, Gautam Tiwari, Parth Suresh, Th{\'e}o Moutakanni, Alejandro
  Castillejo~Munoz, Allen Bolourchi, Pascale Fung, Pinar Donmez, Babak
  Damavandi, Anuj Kumar, and Seungwhan Moon.
\newblock {Plan, Watch, Recover: A Benchmark and Architectures for Proactive
  Procedural Assistance}.
\newblock \emph{arXiv preprint arXiv:2606.04970}, 2026.
\newblock URL \url{https://arxiv.org/abs/2606.04970}.

\bibitem[Kwon et~al.(2026)Kwon, Choi, Kim, Kim, Moon, Kwak, Huang, and
  Yeo]{kwon2025memento}
Taeyoon Kwon, Dongwook Choi, Hyojun Kim, Sunghwan Kim, Seungjun Moon, Beong-woo
  Kwak, Kuan-Hao Huang, and Jinyoung Yeo.
\newblock {Embodied Agents Meet Personalization: Investigating Challenges and
  Solutions Through the Lens of Memory Utilization}.
\newblock In \emph{International Conference on Learning Representations}, 2026.
\newblock URL \url{https://arxiv.org/abs/2505.16348}.

\bibitem[Lei et~al.(2026{\natexlab{a}})Lei, Song, Zhang, Pei, Chen, Yan, Zhao,
  Ding, Zhang, Huang, Wang, Wang, and Li]{lei2026robomemarena}
Huashuo Lei, Wenxuan Song, Huarui Zhang, Jieyuan Pei, Jiayi Chen, Haodong Yan,
  Han Zhao, Pengxiang Ding, Zhipeng Zhang, Lida Huang, Donglin Wang, Yan Wang,
  and Haoang Li.
\newblock {RoboMemArena: A Comprehensive and Challenging Robotic Memory
  Benchmark}.
\newblock \emph{arXiv preprint arXiv:2605.10921}, 2026{\natexlab{a}}.
\newblock URL \url{https://arxiv.org/abs/2605.10921}.

\bibitem[Lei et~al.(2026{\natexlab{b}})Lei, Li, Zhang, Hua, Li, and
  Liu]{lei2026egosat}
Yijia Lei, Jinzhao Li, Yichi Zhang, Jiacheng Hua, Yin Li, and Miao Liu.
\newblock {EgoSAT: A Comprehensive Benchmark of Egocentric Streaming
  Interaction Understanding}.
\newblock In \emph{European Conference on Computer Vision}, 2026{\natexlab{b}}.
\newblock URL \url{https://arxiv.org/abs/2606.24422}.

\bibitem[Mangalam et~al.(2023)Mangalam, Akshulakov, and
  Malik]{mangalam2023egoschema}
Karttikeya Mangalam, Raiymbek Akshulakov, and Jitendra Malik.
\newblock {EgoSchema: A Diagnostic Benchmark for Very Long-form Video Language
  Understanding}.
\newblock In \emph{Advances in Neural Information Processing Systems},
  volume~36, pp.\  46212--46244, 2023.
\newblock \doi{10.52202/075280-2004}.
\newblock URL
  \url{https://proceedings.neurips.cc/paper_files/paper/2023/hash/90ce332aff156b910b002ce4e6880dec-Abstract-Datasets_and_Benchmarks.html}.

\bibitem[{OpenAI}(2026)]{openai2026gpt6sol}
{OpenAI}.
\newblock {Introducing GPT-6 Sol and Luna}.
\newblock Official model release, 2026.
\newblock URL \url{https://openai.com/index/introducing-gpt-6-sol-and-luna/}.
\newblock Accessed September 25, 2026.

\bibitem[P{\u a}tr{\u a}ucean et~al.(2023)P{\u a}tr{\u a}ucean, Smaira, Gupta,
  Recasens~Continente, Markeeva, Banarse, Koppula, Heyward, Malinowski, Yang,
  Doersch, Matejovicova, Sulsky, Miech, Frechette, Klimczak, Koster, Zhang,
  Winkler, Aytar, Osindero, Damen, Zisserman, and
  Carreira]{patraucean2023perceptiontest}
Viorica P{\u a}tr{\u a}ucean, Lucas Smaira, Ankush Gupta, Adri{\`a}
  Recasens~Continente, Larisa Markeeva, Dylan Banarse, Skanda Koppula, Joseph
  Heyward, Mateusz Malinowski, Yi~Yang, Carl Doersch, Tatiana Matejovicova,
  Yury Sulsky, Antoine Miech, Alex Frechette, Hanna Klimczak, Raphael Koster,
  Junlin Zhang, Stephanie Winkler, Yusuf Aytar, Simon Osindero, Dima Damen,
  Andrew Zisserman, and Jo{\~a}o Carreira.
\newblock {Perception Test: A Diagnostic Benchmark for Multimodal Video
  Models}.
\newblock In \emph{Advances in Neural Information Processing Systems},
  volume~36, pp.\  42748--42761, 2023.
\newblock \doi{10.52202/075280-1852}.
\newblock URL
  \url{https://proceedings.neurips.cc/paper_files/paper/2023/hash/8540fba4abdc7f9f7a7b1cc6cd60e409-Abstract.html}.

\bibitem[Peddi et~al.(2024)Peddi, Arya, Challa, Pallapothula, Vyas, Gouripeddi,
  Zhang, Wang, Komaragiri, Ragan, Ruozzi, Xiang, and
  Gogate]{peddi2024captaincook4d}
Rohith Peddi, Shivvrat Arya, Bharath Challa, Likhitha Pallapothula, Akshay
  Vyas, Bhavya Gouripeddi, Qifan Zhang, Jikai Wang, Vasundhara Komaragiri, Eric
  Ragan, Nicholas Ruozzi, Yu~Xiang, and Vibhav Gogate.
\newblock {CaptainCook4D: A Dataset for Understanding Errors in Procedural
  Activities}.
\newblock In \emph{Advances in Neural Information Processing Systems},
  volume~37, pp.\  135626--135679, 2024.
\newblock \doi{10.52202/079017-4307}.
\newblock URL
  \url{https://proceedings.neurips.cc/paper_files/paper/2024/hash/f4a04396c2ed1342a5d8d05e94cb6101-Abstract-Datasets_and_Benchmarks_Track.html}.

\bibitem[{Physical Intelligence} et~al.(2025){Physical Intelligence}, Black,
  Brown, Darpinian, Dhabalia, Driess, Esmail, Equi, Finn, Fusai, Galliker,
  Ghosh, Groom, Hausman, Ichter, Jakubczak, Jones, Ke, LeBlanc, Levine,
  Li-Bell, Mothukuri, Nair, Pertsch, Ren, Shi, Smith, Springenberg, Stachowicz,
  Tanner, Vuong, Walke, Walling, Wang, Yu, and
  Zhilinsky]{physicalintelligence2025pi05}
{Physical Intelligence}, Kevin Black, Noah Brown, James Darpinian, Karan
  Dhabalia, Danny Driess, Adnan Esmail, Michael Equi, Chelsea Finn, Niccolo
  Fusai, Manuel~Y. Galliker, Dibya Ghosh, Lachy Groom, Karol Hausman, Brian
  Ichter, Szymon Jakubczak, Tim Jones, Liyiming Ke, Devin LeBlanc, Sergey
  Levine, Adrian Li-Bell, Mohith Mothukuri, Suraj Nair, Karl Pertsch, Allen~Z.
  Ren, Lucy~Xiaoyang Shi, Laura Smith, Jost~Tobias Springenberg, Kyle
  Stachowicz, James Tanner, Quan Vuong, Homer Walke, Anna Walling, Haohuan
  Wang, Lili Yu, and Ury Zhilinsky.
\newblock {$\pi_{0.5}$: a Vision-Language-Action Model with Open-World
  Generalization}.
\newblock \emph{arXiv preprint arXiv:2504.16054}, 2025.
\newblock URL \url{https://arxiv.org/abs/2504.16054}.

\bibitem[Qiu et~al.(2026)Qiu, Chen, Ge, Ge, Shan, and Liu]{qiu2026egoplan2}
Lu~Qiu, Yi~Chen, Yuying Ge, Yixiao Ge, Ying Shan, and Xihui Liu.
\newblock {EgoPlan-Bench2: A Benchmark for Multimodal Large Language Model
  Planning in Real-World Scenarios}.
\newblock \emph{International Journal of Computer Vision}, 134, 2026.
\newblock \doi{10.1007/s11263-026-02826-y}.
\newblock URL
  \url{https://link.springer.com/article/10.1007/s11263-026-02826-y}.
\newblock Article 222.

\bibitem[{Qwen Team}(2026{\natexlab{a}})]{qwen2026qwen35}
{Qwen Team}.
\newblock {Qwen3.5: Towards Native Multimodal Agents}, February
  2026{\natexlab{a}}.
\newblock URL \url{https://qwen.ai/blog?id=qwen3.5}.

\bibitem[{Qwen Team}(2026{\natexlab{b}})]{qwen2026qwen38}
{Qwen Team}.
\newblock {Qwen3.8-Max: A New Bar for Coding and Cowork}, August
  2026{\natexlab{b}}.
\newblock URL \url{https://qwen.ai/blog?id=qwen3.8}.

\bibitem[Sermanet et~al.(2024)Sermanet, Ding, Zhao, Xia, Dwibedi,
  Gopalakrishnan, Chan, Dulac-Arnold, Maddineni, Joshi, Florence, Han, Baruch,
  Lu, Mirchandani, Xu, Sanketi, Hausman, Shafran, Ichter, and
  Cao]{sermanet2024robovqa}
Pierre Sermanet, Tianli Ding, Jeffrey Zhao, Fei Xia, Debidatta Dwibedi,
  Keerthana Gopalakrishnan, Christine Chan, Gabriel Dulac-Arnold, Sharath
  Maddineni, Nikhil~J Joshi, Pete Florence, Wei Han, Robert Baruch, Yao Lu,
  Suvir Mirchandani, Peng Xu, Pannag Sanketi, Karol Hausman, Izhak Shafran,
  Brian Ichter, and Yuan Cao.
\newblock {RoboVQA: Multimodal Long-Horizon Reasoning for Robotics}.
\newblock In \emph{IEEE International Conference on Robotics and Automation},
  pp.\  645--652, 2024.
\newblock \doi{10.1109/ICRA57147.2024.10610216}.
\newblock URL \url{https://arxiv.org/abs/2311.00899}.

\bibitem[Wang et~al.(2023)Wang, Kwon, Rad, Pan, Chakraborty, Andrist, Bohus,
  Feniello, Tekin, Frujeri, Joshi, and Pollefeys]{wang2023holoassist}
Xin Wang, Taein Kwon, Mahdi Rad, Bowen Pan, Ishani Chakraborty, Sean Andrist,
  Dan Bohus, Ashley Feniello, Bugra Tekin, Felipe~Vieira Frujeri, Neel Joshi,
  and Marc Pollefeys.
\newblock {HoloAssist: an Egocentric Human Interaction Dataset for Interactive
  AI Assistants in the Real World}.
\newblock In \emph{Proceedings of the IEEE/CVF International Conference on
  Computer Vision}, pp.\  20270--20281, 2023.
\newblock URL
  \url{https://openaccess.thecvf.com/content/ICCV2023/html/Wang_HoloAssist_an_Egocentric_Human_Interaction_Dataset_for_Interactive_AI_Assistants_ICCV_2023_paper.html}.

\bibitem[Wang et~al.(2026)Wang, Zhang, Yu, Zhang, Zhao, Yoon, Lee, Bertasius,
  and Bansal]{wang2026egomemreason}
Ziyang Wang, Yue Zhang, Shoubin Yu, Ce~Zhang, Zengqi Zhao, Jaehong Yoon, Hyunji
  Lee, Gedas Bertasius, and Mohit Bansal.
\newblock {EgoMemReason: A Memory-Driven Reasoning Benchmark for Long-Horizon
  Egocentric Video Understanding}.
\newblock In \emph{Conference on Language Modeling}, 2026.
\newblock URL \url{https://arxiv.org/abs/2605.09874v2}.

\bibitem[Wei et~al.(2022)Wei, Wang, Schuurmans, Bosma, Ichter, Xia, Chi, Le,
  and Zhou]{wei2022chain}
Jason Wei, Xuezhi Wang, Dale Schuurmans, Maarten Bosma, Brian Ichter, Fei Xia,
  Ed~Chi, Quoc~V. Le, and Denny Zhou.
\newblock {Chain-of-Thought} prompting elicits reasoning in large language
  models.
\newblock In \emph{Advances in Neural Information Processing Systems},
  volume~35, pp.\  24824--24837, 2022.
\newblock \doi{10.52202/068431-1800}.
\newblock URL
  \url{https://proceedings.neurips.cc/paper/2022/hash/9d5609613524ecf4f15af0f7b31abca4-Abstract-Conference.html}.

\bibitem[Wu et~al.(2024)Wu, Li, Chen, and Li]{wu2024longvideobench}
Haoning Wu, Dongxu Li, Bei Chen, and Junnan Li.
\newblock {LongVideoBench: A Benchmark for Long-context Interleaved
  Video-Language Understanding}.
\newblock In \emph{Advances in Neural Information Processing Systems},
  volume~37, pp.\  28828--28857, 2024.
\newblock \doi{10.52202/079017-0907}.
\newblock URL
  \url{https://papers.nips.cc/paper_files/paper/2024/hash/329ad516cf7a6ac306f29882e9c77558-Abstract-Datasets_and_Benchmarks_Track.html}.

\bibitem[{Xiaomi MiMo Team}(2026)]{xiaomi2026mimov26}
{Xiaomi MiMo Team}.
\newblock {MiMo-V2.6: Scaling Up Reinforcement Learning for Self-Improvement}.
\newblock Official model release, September 2026.
\newblock URL \url{https://mimo.mi.com/docs/en-US/news/latest/v2-6}.
\newblock Accessed September 25, 2026.

\bibitem[Yadav et~al.(2026)Yadav, Ali, Gupta, Gal, and
  Kira]{yadav2026findingdory}
Karmesh Yadav, Yusuf Ali, Gunshi Gupta, Yarin Gal, and Zsolt Kira.
\newblock {FindingDory: A Benchmark to Evaluate Memory in Embodied Agents}.
\newblock In \emph{European Conference on Computer Vision}, 2026.
\newblock URL \url{https://arxiv.org/abs/2506.15635}.

\bibitem[Yan et~al.(2025)Yan, Ren, Liu, Xu, Wang, Wang, Zhong, Wang, Zhang,
  Chen, Sun, Luo, Zhang, Sun, Zhang, and Li]{yan2025teleego}
Jiaqi Yan, Ruilong Ren, Jingren Liu, Shuning Xu, Ling Wang, Yiheng Wang, Xinlin
  Zhong, Yun Wang, Long Zhang, Xiangyu Chen, Changzhi Sun, Jixiang Luo, Dell
  Zhang, Hao Sun, Chi Zhang, and Xuelong Li.
\newblock {TeleEgo: Benchmarking Egocentric AI Assistants in the Wild}.
\newblock \emph{arXiv preprint arXiv:2510.23981}, 2025.
\newblock URL \url{https://arxiv.org/abs/2510.23981}.

\bibitem[Yang et~al.(2025{\natexlab{a}})Yang, Liu, Guo, Dong, Zhang, Zhang,
  Wang, Zhou, Xie, Wang, Ouyang, Lin, Cominelli, Cai, Li, Zhang, Zhang, Hong,
  Widmer, Gringoli, Yang, and Liu]{yang2025egolife}
Jingkang Yang, Shuai Liu, Hongming Guo, Yuhao Dong, Xiamengwei Zhang, Sicheng
  Zhang, Pengyun Wang, Zitang Zhou, Binzhu Xie, Ziyue Wang, Bei Ouyang, Zhengyu
  Lin, Marco Cominelli, Zhongang Cai, Bo~Li, Yuanhan Zhang, Peiyuan Zhang,
  Fangzhou Hong, Joerg Widmer, Francesco Gringoli, Lei Yang, and Ziwei Liu.
\newblock {EgoLife: Towards Egocentric Life Assistant}.
\newblock In \emph{Proceedings of the IEEE/CVF Conference on Computer Vision
  and Pattern Recognition}, pp.\  28885--28900, 2025{\natexlab{a}}.
\newblock URL
  \url{https://openaccess.thecvf.com/content/CVPR2025/html/Yang_EgoLife_Towards_Egocentric_Life_Assistant_CVPR_2025_paper.html}.

\bibitem[Yang et~al.(2025{\natexlab{b}})Yang, Chen, Zhang, Zhao, Qian, Wang,
  Wang, Koripella, Movahedi, Li, Ji, Zhang, and Zhang]{yang2025embodiedbench}
Rui Yang, Hanyang Chen, Junyu Zhang, Mark Zhao, Cheng Qian, Kangrui Wang,
  Qineng Wang, Teja~Venkat Koripella, Marziyeh Movahedi, Manling Li, Heng Ji,
  Huan Zhang, and Tong Zhang.
\newblock {EmbodiedBench: Comprehensive Benchmarking Multi-modal Large Language
  Models for Vision-Driven Embodied Agents}.
\newblock In \emph{Proceedings of the 42nd International Conference on Machine
  Learning}, volume 267 of \emph{Proceedings of Machine Learning Research},
  pp.\  70576--70631, 2025{\natexlab{b}}.
\newblock URL \url{https://proceedings.mlr.press/v267/yang25f.html}.

\bibitem[Ye et~al.(2025)Ye, Zhang, Daxberger, Chen, Lin, Li, Zhang, You, Xu,
  Gan, Lu, and Yang]{ye2025mmego}
Hanrong Ye, Haotian Zhang, Erik Daxberger, Lin Chen, Zongyu Lin, Yanghao Li,
  Bowen Zhang, Haoxuan You, Dan Xu, Zhe Gan, Jiasen Lu, and Yinfei Yang.
\newblock {MMEgo: Towards Building Egocentric Multimodal LLMs for Video QA}.
\newblock In \emph{International Conference on Learning Representations}, 2025.
\newblock URL
  \url{https://proceedings.iclr.cc/paper_files/paper/2025/hash/b29a95e7d9f1e1a6dfc567b556733744-Abstract-Conference.html}.

\bibitem[Yuan et~al.(2025)Yuan, Dang, Li, Li, Jiao, Li, Zhao, Wang, Zhang,
  Xiao, and Zhuang]{yuan2025eocbench}
Yuqian Yuan, Ronghao Dang, Long Li, Wentong Li, Dian Jiao, Xin Li, Deli Zhao,
  Fan Wang, Wenqiao Zhang, Jun Xiao, and Yueting Zhuang.
\newblock {EOC-Bench: Can MLLMs Identify, Recall, and Forecast Objects in an
  Egocentric World?}
\newblock In \emph{Advances in Neural Information Processing Systems},
  volume~38, 2025.
\newblock \doi{10.52202/085713-4489}.
\newblock URL
  \url{https://papers.neurips.cc/paper_files/paper/2025/hash/c398086709acb9938fb8c4e0868e1643-Abstract-Datasets_and_Benchmarks_Track.html}.

\bibitem[Zhang et~al.(2025{\natexlab{a}})Zhang, Wang, Gu, Hamidizadeh, Zhang,
  Liu, Wang, Bravo, Dong, Zhou, Cao, Quan, Zhuang, Zhang, and
  Hao]{zhang2025etplan}
Lingfeng Zhang, Yuening Wang, Hongjian Gu, Atia Hamidizadeh, Zhanguang Zhang,
  Yuecheng Liu, Yutong Wang, David Gamaliel~Arcos Bravo, Junyi Dong, Shunbo
  Zhou, Tongtong Cao, Xingyue Quan, Yuzheng Zhuang, Yingxue Zhang, and Jianye
  Hao.
\newblock {ET-Plan-Bench: Embodied Task-Level Planning Benchmark Towards
  Spatial-Temporal Cognition with Foundation Models}.
\newblock In \emph{IEEE/RSJ International Conference on Intelligent Robots and
  Systems}, pp.\  21566--21573, 2025{\natexlab{a}}.
\newblock \doi{10.1109/IROS60139.2025.11246658}.
\newblock URL \url{https://arxiv.org/abs/2410.14682}.

\bibitem[Zhang et~al.(2025{\natexlab{b}})Zhang, Dong, Lin, Madotto, Kumar,
  Damavandi, Chai, and Moon]{zhang2025proassist}
Yichi Zhang, Xin~Luna Dong, Zhaojiang Lin, Andrea Madotto, Anuj Kumar, Babak
  Damavandi, Joyce Chai, and Seungwhan Moon.
\newblock {Proactive Assistant Dialogue Generation from Streaming Egocentric
  Videos}.
\newblock In \emph{Proceedings of the 2025 Conference on Empirical Methods in
  Natural Language Processing}, pp.\  12044--12068. Association for
  Computational Linguistics, 2025{\natexlab{b}}.
\newblock \doi{10.18653/v1/2025.emnlp-main.605}.
\newblock URL \url{https://aclanthology.org/2025.emnlp-main.605/}.

\bibitem[Zhou et~al.(2025)Zhou, Cao, Zheng, Liu, Zheng, Liu, Kristensson,
  Mayol-Cuevas, Zhang, Lin, and Shen]{zhou2025xlebench}
Wenqi Zhou, Kai Cao, Hao Zheng, Yunze Liu, Xinyi Zheng, Miao Liu, Per~Ola
  Kristensson, Walterio~W. Mayol-Cuevas, Fan Zhang, Weizhe Lin, and Junxiao
  Shen.
\newblock {X-LeBench: A Benchmark for Extremely Long Egocentric Video
  Understanding}.
\newblock In \emph{Findings of the Association for Computational Linguistics:
  EMNLP 2025}, pp.\  15206--15222. Association for Computational Linguistics,
  2025.
\newblock \doi{10.18653/v1/2025.findings-emnlp.822}.
\newblock URL \url{https://aclanthology.org/2025.findings-emnlp.822/}.

\end{thebibliography}
\bibliographystyle{ember_article}

\appendix
\section{Action with Anomalies Annotation Schema}
\label{app:action_annotation}
For samples involving anomalies (L2--L4), we use four labels to organize abnormal causes and consequences into causal chains, which is a key design of this benchmark. We need abnormal situations and their repaying actions to construct bidirectional memory-causal questions. Therefore, we adopt a unified structured annotation to record the correlation between event stages, obligations, abnormal causes, and consequences using four labels (shown in Table~\ref{tab:action_annotation_schema}): \textit{execution\_stage}, \textit{obligation\_related}, \textit{accident\_cause}, and \textit{accident\_consequence}. The \textit{execution\_stage} assigns the labels of ``before'', ``during'', ``repaying'' or ``after'' to each action on the corresponding event timeline, indicating the abnormal stage it is in. The labels \textit{accident\_cause} and \textit{accident\_consequence} describe, respectively, the specific events or actions that trigger a change in an object's state during a given period and the effects of that change on ongoing task execution or the objects involved; both fields are filled in only when \textit{execution\_stage} is ``during'' (i.e., when the anomaly occurs during the execution of the corresponding subtask). Additionally, \textit{obligation\_related} should be filled in during the ``repaying'' stage, indicating the restoration or correction matters corresponding to the current processing action, containing the cause and consequences of the repaying abnormal. Except for \textit{execution\_stage}, we set all other fields to ``false'' in stages where they do not apply. Here ``false'' indicates that this field is not filled in for the current action instead of meaning that the effects of previous events have disappeared.
\begin{table}[H]
\centering
\caption{Action-level annotation fields. Each action receives an execution-stage label. The other fields are populated only at the specified stage and are set to the Boolean value \texttt{false} otherwise.}
\label{tab:action_annotation_schema}
\small
\setlength{\tabcolsep}{4pt}
\renewcommand{\arraystretch}{1.15}
\begin{tabularx}{\linewidth}{@{}>{\raggedright\arraybackslash}p{0.23\linewidth} *{3}{>{\raggedright\arraybackslash}X}@{}}
\toprule
Field & Guiding question & When to populate & Example values \\
\midrule

\texttt{execution\_\allowbreak stage}
& Where does the action occur in the accident timeline?
& Every action.
& \texttt{before}, \texttt{during}, \texttt{repaying}, \texttt{after}. \\
\addlinespace

\texttt{obligation\_\allowbreak related}
& Which accident does the action help remedy?
& Only in the \texttt{repaying} stage; \texttt{false} otherwise.
& ``Replenish the milk'' or \texttt{false}. \\
\addlinespace

\texttt{accident\_\allowbreak cause}
& Which action caused the accident?
& Only in the \texttt{during} stage; \texttt{false} otherwise.
& ``Turning around caused the insulated cup to hit the counter edge.'' \\
\addlinespace

\texttt{accident\_\allowbreak consequence}
& What consequences did the accident cause?
& Only in the \texttt{during} stage; \texttt{false} otherwise.
&``Cup damage''; ``Raw materials contaminated'', etc. \\

\bottomrule
\end{tabularx}
\end{table}
\begin{table}[H]
\centering
\caption{Example execution-stage annotations for a sequence of food-preparation actions. The \texttt{after} stage begins only when all remedial actions have been completed. } 
\label{tab:execution_stage_example}
\small
\setlength{\tabcolsep}{4pt}
\renewcommand{\arraystretch}{1.15}
\begin{tabularx}{\linewidth}{@{}>{\raggedright\arraybackslash}p{0.40\linewidth} >{\raggedright\arraybackslash}p{0.22\linewidth} >{\raggedright\arraybackslash}X@{}}
\toprule
Action & \texttt{execution\_\allowbreak stage} & Explanation \\
\midrule

Cut cheese normally.
& \texttt{before}
& No accident has occurred yet. \\
\addlinespace

Slice a cucumber (accidentally knocking some slices onto the floor).
& \texttt{during}
& The first accident occurs. \\
\addlinespace

Pick up the fallen cucumber slices.
& \texttt{repaying}
& Attempt to address the first accident. \\
\addlinespace

Throw the cucumber slices into the trash bin.
& \texttt{repaying}
& --- \\
\addlinespace

Get a cloth, wipe up the water, and put the cloth back.
& \texttt{repaying}
& Fulfill the floor-cleaning obligation. \\
\addlinespace

Go to the refrigerator, get a fresh cucumber, and slice it again.
& \texttt{repaying}
& Fulfill the obligation to redo the preparation. \\
\addlinespace

Resume normal food preparation after replacing the lost slices.
& \texttt{after}
& All remedial actions have been completed. \\

\bottomrule
\end{tabularx}
\end{table}
Here, the action sequence is just for demonstration. In our real annotations, each subtask only contains a single atomic action.
\paragraph{Constructing paired causal traceback questions.} For eligible L2--L4 prediction questions, we start from the recovery subtask labeled \textit{execution\_stage} = ``repaying'' and use its \textit{obligation\_related} annotation to locate the corresponding anomalous subtask labeled \textit{execution\_stage} = ``during''. We then retrieve the \textit{accident\_cause} and \textit{accident\_consequence} annotations of that subtask to establish the historical basis for the reference next action. The corresponding historical event serves as the reference answer to the paired causal traceback question. Prediction questions without a uniquely traceable event remain unpaired.

\section{Evaluation Prompts}
\label{app:prompts}

This appendix lists the prompt templates used in the evaluations. Fields in braces are instantiated for each question; the remaining prompt text is reproduced verbatim. Box titles and the surrounding descriptions are explanatory labels and are not part of the model input.

\subsection{Input descriptions}
\label{app:prompt_input_descriptions}

For main next-action prediction, the following texts precede the history video and the current frame, respectively. The same video description is used in the explicit-history ablations.

\begin{emberpromptcontext}{Prediction video preamble}
\noindent History video at 5 fps, uniformly sampled from the start up to the decision moment; then the current frame at the decision moment.\par
\end{emberpromptcontext}

\begin{emberpromptcontext}{Prediction current-frame preamble}
\noindent Current frame at the decision moment:\par
\end{emberpromptcontext}

For causal traceback, the following text precedes the history video.

\begin{emberpromptcontext}{Causal-traceback video preamble}
\noindent History video at 5 fps with OPTION A\textendash{}D labels, ending before the known action; then the current frame at the decision moment.\par
\end{emberpromptcontext}

\subsection{Main evaluation: video input (V)}
\label{app:prompt_main}

The main next-action prediction ($P$) and causal-traceback ($C$) templates are shown below. For causal traceback, OPTION A\textendash D labels are overlaid on the history video to identify candidate intervals; the prompt does not describe the contents of these intervals.

\begin{emberprompt}[label={prompt:main_prediction}]{Next-action prediction ($V$, $P$)}
\noindent Task instruction: \texttt{\{objective\}}\par
\noindent Question: What is the most likely next action to perform?\par
\noindent A. \texttt{\{option{\ttfamily\char95}A\}}\par
\noindent B. \texttt{\{option{\ttfamily\char95}B\}}\par
\noindent C. \texttt{\{option{\ttfamily\char95}C\}}\par
\noindent D. \texttt{\{option{\ttfamily\char95}D\}}\par
\vspace{\baselineskip}
\vspace{\baselineskip}
\noindent The history video is a uniformly sampled visual prefix up to the decision moment, followed by the current decision frame. Use both the video and the current frame.\par
\noindent If an action log is provided, it contains only completed subtasks before the decision and does not include privileged anomaly fields.\par
\noindent Do not write a chain of thought, explanation, or markdown.\par
\noindent Output exactly one line and nothing else:\par
\noindent Final Answer: [A/B/C/D]\par
\end{emberprompt}

\begin{emberprompt}[label={prompt:main_traceback}]{Causal traceback ($V$, $C$)}
\noindent Task instruction: \texttt{\{objective\}}\par
\noindent The next action is: \texttt{\char34}\texttt{\{known{\ttfamily\char95}action\}}\texttt{\char34}\par
\noindent Question: Which marked historical segment contains the core incident that made this next action necessary?\par
\noindent A. The event shown while OPTION A appears in the history video.\par
\noindent B. The event shown while OPTION B appears in the history video.\par
\noindent C. The event shown while OPTION C appears in the history video.\par
\noindent D. The event shown while OPTION D appears in the history video.\par
\vspace{\baselineskip}
\vspace{\baselineskip}
\noindent Each on-screen label OPTION A\textendash{}D identifies one candidate historical interval.\par
\noindent Select the marked segment that contains the core incident that made this next action necessary.\par
\noindent Do not write a chain of thought, explanation, or markdown.\par
\noindent Output exactly one line and nothing else:\par
\noindent Final Answer: [A/B/C/D]\par
\end{emberprompt}

\subsection{Explicit-history ablations}
\label{app:prompt_history_ablation}

These ablations use 151 next-action prediction questions. The $V$ condition reuses the main-evaluation answers. Both ablation conditions use the prediction video preamble above.

The field \texttt{\{action\_log\_json\}} is a JSON list of subtasks completed before the decision. Its fields are \texttt{subtask}, \texttt{action}, \texttt{start\_frame}, \texttt{stop\_frame}, \texttt{start\_sec}, \texttt{stop\_sec}, \texttt{same\_object}, \texttt{left\_hand}, \texttt{right\_hand}, and \texttt{navigation}. It excludes anomaly-stage, obligation, cause, and consequence fields.

\begin{emberprompt}[label={prompt:action_log}]{Action log ($V{+}L$)}
\noindent Task instruction: \texttt{\{objective\}}\par
\noindent Question: What is the most likely next action to perform?\par
\noindent A. \texttt{\{option{\ttfamily\char95}A\}}\par
\noindent B. \texttt{\{option{\ttfamily\char95}B\}}\par
\noindent C. \texttt{\{option{\ttfamily\char95}C\}}\par
\noindent D. \texttt{\{option{\ttfamily\char95}D\}}\par
\vspace{\baselineskip}
\vspace{\baselineskip}
\noindent You are given the complete non-privileged action log of completed subtasks up to the current decision frame. Anomaly stage, obligation, cause, and consequence fields are omitted.\par
\noindent action{\ttfamily\char95}log{\ttfamily\char95}nonprivileged:\par
\noindent \texttt{\{action{\ttfamily\char95}log{\ttfamily\char95}json\}}\par
\vspace{\baselineskip}
\noindent The history video is a uniformly sampled visual prefix up to the decision moment, followed by the current decision frame. Use both the video and the current frame.\par
\noindent If an action log is provided, it contains only completed subtasks before the decision and does not include privileged anomaly fields.\par
\noindent Do not write a chain of thought, explanation, or markdown.\par
\noindent Output exactly one line and nothing else:\par
\noindent Final Answer: [A/B/C/D]\par
\end{emberprompt}

\begin{emberprompt}[label={prompt:incident_annotations}]{Action log and incident annotations ($V{+}L{+}E$)}
\noindent Task instruction: \texttt{\{objective\}}\par
\noindent Question: What is the most likely next action to perform?\par
\noindent A. \texttt{\{option{\ttfamily\char95}A\}}\par
\noindent B. \texttt{\{option{\ttfamily\char95}B\}}\par
\noindent C. \texttt{\{option{\ttfamily\char95}C\}}\par
\noindent D. \texttt{\{option{\ttfamily\char95}D\}}\par
\vspace{\baselineskip}
\vspace{\baselineskip}
\noindent You are given the complete non-privileged action log of completed subtasks up to the current decision frame. Anomaly stage and obligation fields are omitted. Annotated accident cause and consequence for incidents in these completed subtasks are listed after the log.\par
\noindent action{\ttfamily\char95}log{\ttfamily\char95}nonprivileged:\par
\noindent \texttt{\{action{\ttfamily\char95}log{\ttfamily\char95}json\}}\par
\vspace{\baselineskip}
\noindent incident{\ttfamily\char95}annotations (accident cause and consequence):\par
\noindent [\par
\noindent \hspace*{1.0em}\{\par
\noindent \hspace*{2.0em}\texttt{\char34}subtask\texttt{\char34}: \texttt{\char34}\texttt{\{k\}}\texttt{\char34},\par
\noindent \hspace*{2.0em}\texttt{\char34}action\texttt{\char34}: \texttt{\char34}\texttt{\{subtask{\ttfamily\char95}action\}}\texttt{\char34},\par
\noindent \hspace*{2.0em}\texttt{\char34}accident{\ttfamily\char95}cause\texttt{\char34}: \texttt{\char34}\texttt{\{accident{\ttfamily\char95}cause\}}\texttt{\char34},\par
\noindent \hspace*{2.0em}\texttt{\char34}accident{\ttfamily\char95}consequence\texttt{\char34}: \texttt{\char34}\texttt{\{accident{\ttfamily\char95}consequence\}}\texttt{\char34}\par
\noindent \hspace*{1.0em}\}\par
\noindent ]\par
\vspace{\baselineskip}
\noindent The history video is a uniformly sampled visual prefix up to the decision moment, followed by the current decision frame. Use both the video and the current frame.\par
\noindent The action log contains only completed subtasks before the decision. Incident annotations (accident cause and consequence) refer only to those completed subtasks; they do not state the next action.\par
\noindent Do not write a chain of thought, explanation, or markdown.\par
\noindent Output exactly one line and nothing else:\par
\noindent Final Answer: [A/B/C/D]\par
\end{emberprompt}

Each annotated incident occupies one list item. Missing fields are filled with \texttt{(not annotated)}. When there are no annotated incidents before the decision, the incident-annotation block is replaced by the following text.

\begin{emberpromptcontext}{Replacement for an empty incident-annotation list}
\noindent incident{\ttfamily\char95}annotations (accident cause and consequence):\par
\noindent []  (no incident was annotated in the completed subtasks)\par
\end{emberpromptcontext}

\subsection{Simple causal chain-of-thought guidance}
\label{app:prompt_causal_cot}

This experiment uses 151 next-action prediction questions. The direct-answering control uses Prompt~\ref{prompt:main_prediction}. The CoT condition replaces its final three output-instruction lines with the assessment and two-line output requirements below; the complete resulting template is provided for clarity.

\begin{emberprompt}[label={prompt:causal_cot}]{Causal CoT for next-action prediction}
\noindent Task instruction: \texttt{\{objective\}}\par
\noindent Question: What is the most likely next action to perform?\par
\noindent A. \texttt{\{option{\ttfamily\char95}A\}}\par
\noindent B. \texttt{\{option{\ttfamily\char95}B\}}\par
\noindent C. \texttt{\{option{\ttfamily\char95}C\}}\par
\noindent D. \texttt{\{option{\ttfamily\char95}D\}}\par
\vspace{\baselineskip}
\vspace{\baselineskip}
\noindent The history video is a uniformly sampled visual prefix up to the decision moment, followed by the current decision frame. Use both the video and the current frame.\par
\noindent If an action log is provided, it contains only completed subtasks before the decision and does not include privileged anomaly fields.\par
\noindent Before choosing, briefly state whether any observed event disrupted task execution and what effect, if any, still matters for the current decision. Do not assume a disruption occurred or remains unresolved. Then select the next action.\par
\noindent Output exactly two lines, without markdown:\par
\noindent Assessment: \texttt{\textless{}at most two sentences and 60 words\textgreater{}}\par
\noindent Final Answer: [A/B/C/D]\par
\end{emberprompt}

\subsection{Current-frame-only control}
\label{app:prompt_current_frame_only}

This control uses 548 next-action prediction questions. No history video or video preamble is provided; only the decision frame and the following text are sent to the model.

\begin{emberprompt}[label={prompt:current_frame_only}]{Current-frame-only next-action prediction}
\noindent Task instruction: \texttt{\{objective\}}\par
\noindent Question: What is the most likely next action to perform?\par
\noindent A. \texttt{\{option{\ttfamily\char95}A\}}\par
\noindent B. \texttt{\{option{\ttfamily\char95}B\}}\par
\noindent C. \texttt{\{option{\ttfamily\char95}C\}}\par
\noindent D. \texttt{\{option{\ttfamily\char95}D\}}\par
\vspace{\baselineskip}
\noindent Choose the best next action based on the task and the current image.\par
\noindent Output exactly one line: Final Answer: [A/B/C/D]\par
\end{emberprompt}

\subsection{Answer parsing}
\label{app:prompt_answer_parsing}

All evaluations use the same parsing rule. The answer letter following \texttt{Final Answer:} is extracted, allowing square brackets, bold formatting, or heading markers. If this marker occurs more than once with inconsistent answer letters, the response is invalid. If the marker is absent, a response is accepted only when its final line contains a single letter from A to D. A question is counted as incorrect if no valid answer is obtained after technical retries.

\section{Baseline Models}
\label{app:baseline_models}
 We evaluate 16 multimodal large language models (MLLMs) on EMBER-Bench,
covering seven closed-source models and nine open-source models. Closed-source models are accessible only via APIs, whereas open-source models publicly release their weights.

\paragraph{Closed-source models.}
We evaluate
\href{https://ai.google.dev/gemini-api/docs/models/gemini-3.8-flash}{Gemini 3.8 Flash}
\citep{google2026gemini38flash};
\href{https://ai.google.dev/gemini-api/docs/models/gemini-3.1-pro-preview}{Gemini 3.1 Pro Preview}
\citep{google2026gemini31pro};
\href{https://seed.bytedance.com/en/blog/seed2-1-officially-released-advancing-ai-productivity}{Doubao-Seed-2.1-Pro}
\citep{bytedance2026seed21};
\href{https://qwen.ai/blog?id=qwen3.8}{Qwen3.8 Max}
\citep{qwen2026qwen38};
\href{https://developers.openai.com/api/docs/models/gpt-6-sol}{GPT-6 Sol}
\citep{openai2026gpt6sol};
\href{https://www.alibabacloud.com/help/en/model-studio/qwen3-8-omni-flash}{Qwen3.8 Omni Flash}
\citep{alibaba2026qwen38omniflash}; and
\href{https://www.alibabacloud.com/help/en/model-studio/qwen3-8-flash}{Qwen3.8 Flash}
\citep{alibaba2026qwen38flash}.

\paragraph{Open-source models.}
We evaluate three Qwen3.5 variants:
\href{https://huggingface.co/Qwen/Qwen3.5-397B-A17B}{Qwen3.5 397B-A17B},
\href{https://huggingface.co/Qwen/Qwen3.5-35B-A3B}{Qwen3.5 35B-A3B}, and
\href{https://huggingface.co/Qwen/Qwen3.5-122B-A10B}{Qwen3.5 122B-A10B}
\citep{qwen2026qwen35}.
We additionally evaluate
\href{https://huggingface.co/Qwen/Qwen3.8-27B}{Qwen3.8 27B}
\citep{qwen2026qwen38};
\href{https://mimo.mi.com/models/en-US/mimo-v2.6-pro}{MiMo V2.6 Pro}
and
\href{https://mimo.mi.com/models/en-US/mimo-v2.6-flash}{MiMo V2.6 Flash}
\citep{xiaomi2026mimov26}; and
\href{https://github.com/QwenLM/Qwen3-VL}{Qwen3-VL 235B-A22B}, \href{https://github.com/QwenLM/Qwen3-VL}{Qwen3-VL 32B},
and
\href{https://github.com/QwenLM/Qwen3-VL}{Qwen3-VL 8B}
\citep{bai2025qwen3vl}.
\paragraph{API model versions.}
Some API evaluations used model aliases without an exposed immutable
snapshot. Their results therefore correspond to the service versions
available during the reported evaluation periods.
\section{Validation Test for Next-Action Prediction Questions}
\label{app:validation_test}
\begin{table}[H]
\centering
\caption{Model accuracy under the setting with goal, question, options, and the current frame.}
\small
\setlength{\tabcolsep}{10pt}
\renewcommand{\arraystretch}{1.12}
\begin{tabular}{@{}lrr@{}}
\toprule
Model & Correct / 548 & Accuracy (\%) \\
\midrule
Gemini 3.8 Flash       & \textbf{166} & \textbf{30.29} \\
Gemini 3.1 Pro         & 147 & 26.82 \\
Qwen3.8 Omni Flash     & 146 & 26.64 \\
Qwen3.8 Max            & 146 & 26.64 \\
Qwen3-VL 32B           & 145 & 26.46 \\
Qwen3.8 Flash          & 143 & 26.09 \\
MiMo V2.6 Pro          & 144 & 26.28 \\
Qwen3.5 122B-A10B      & 141 & 25.73 \\
Qwen3.5 397B-A17B      & 143 & 26.09 \\
Qwen3.8 27B            & 142 & 25.91 \\
Qwen3-VL 8B            & 134 & 24.45 \\
Qwen3.5 35B-A3B        & 152 & 27.74 \\
Qwen3-VL 235B-A22B     & 145 & 26.46 \\
GPT-6 Sol             & 141 & 25.73 \\
MiMo V2.6 Flash        & 138 & 25.18 \\
Doubao-Seed-2.1-Pro     & 150 & 27.37 \\
\midrule
Mean (16 models)      & 145.19 & 26.49 \\
\bottomrule
\end{tabular}
\end{table}
\section{Additional Experimental Details}
\label{app:experiments}

\subsection{Evaluation Protocol}
\label{app:exp_protocol}
All 16 models follow the same evaluation protocol:
\begin{itemize}
    \item \textbf{Calls.} Each question is answered in a single stateless, single-turn API call; no context is shared across questions. The temperature is set to 0 when the API supports it.
    \item \textbf{Reasoning.} Reasoning is disabled when the API supports it; otherwise, the lowest level is used.
    \item \textbf{Video input.} The history video is uniformly sampled at 5\,Hz from the start of the task to the decision time and is followed by the current decision frame. For models with smaller frame budgets, the sequence is proportionally downsampled so that the entire history fits without temporal truncation; every model thus sees an evenly spaced sequence covering the full decision history. No audio is used. Reported results reflect each model's best achievable performance under its own interface constraints.
    \item \textbf{Prompts.} All models use the same prompts for each question type; the full prompts are given in \hyperref[app:prompts]{Appendix~\ref*{app:prompts}}.
    \item \textbf{Retries and scoring.} Failed requests are retried; questions without a valid answer after retries are scored incorrect.
\end{itemize}

\subsection{Analysis Samples and Statistical Method}
\label{app:stats}
Table~\ref{tab:app_cohorts} lists the models and questions used in each analysis. Each $C$ question shares its task and decision frame with one $P$ question, which yields the 151 matched pairs.

\begin{table}[H]
\centering
\caption{Samples used in each analysis.}
\label{tab:app_cohorts}
\small
\begin{tabular}{lrl}
\toprule
Analysis & Models & Questions / responses \\
\midrule
Main leaderboard (Table~\ref{tab:main}) & 16 & 699 per model (548 $P$, 151 $C$) \\
Temporal analysis (Figure~\ref{fig:exp_analysis}(a--b)) & 16 & 151 matched $P$--$C$ pairs \\
Information ablation (Figure~\ref{fig:exp_analysis}(c--d)) & 6 & 151 $P$ questions $\times$ 3 inputs \\
Joint analysis (Figure~\ref{fig:exp_analysis}(e--f)) & 16 & 151 matched pairs; 2,416 paired responses \\
Causal CoT (Table~\ref{tab:exp_cot}) & 3 & The same 151 $P$ questions \\
\bottomrule
\end{tabular}
\end{table}

All confidence intervals use a task-clustered bootstrap: tasks are resampled with replacement, all questions and model responses of a sampled task are drawn together, and the interval is given by the 2.5th and 97.5th percentiles of the bootstrap distribution.

\subsection{Temporal Analysis}
\label{app:temporal}
Total history duration is the time from the start of the video to the decision frame; the event-to-decision interval is the time from the end of the correct clip of the $C$ question to the decision frame. Each variable is split into tertiles over the 151 pairs, and tied values at a boundary are assigned to the same group (Table~\ref{tab:app_tertiles}).

\begin{table}[H]
\centering
\caption{Observed ranges of the temporal tertiles.}
\label{tab:app_tertiles}
\small
\begin{tabular}{llr}
\toprule
Variable & Group & Range (s) \\
\midrule
\multirow{3}{*}{Total history duration} & Short & 21.17--83.50 \\
 & Medium & 84.57--134.63 \\
 & Long & 136.60--594.80 \\
\midrule
\multirow{3}{*}{Event-to-decision interval} & Short & 0--7.97 \\
 & Medium & 8.03--30.37 \\
 & Long & 30.57--163.03 \\
\bottomrule
\end{tabular}
\end{table}

The supplementary regression takes question-level accuracy as the dependent variable and jointly includes $\log_2(\text{total history duration})$, $\log_2(\text{interval}+1)$, and level indicators. Each coefficient is the change in accuracy (percentage points) per doubling of the variable; for the interval, a doubling refers to the interval plus 1\,s. The results indicate associations only and are not interpreted causally (Table~\ref{tab:app_regression}).

\begin{table}[H]
\centering
\caption{Temporal associations after controlling for level.}
\label{tab:app_regression}
\small
\begin{tabular}{llrr}
\toprule
Outcome & Predictor (per doubling) & Coefficient (pp) & 95\% CI \\
\midrule
$P$ & Total history duration & $+11.7$ & [$+5.5$, $+17.5$] \\
$P$ & Interval $+$ 1\,s & $+1.7$ & [$-0.6$, $+4.4$] \\
$C$ & Total history duration & $+3.9$ & [$-4.0$, $+11.9$] \\
$C$ & Interval $+$ 1\,s & $-3.7$ & [$-6.7$, $-0.9$] \\
\bottomrule
\end{tabular}
\end{table}

\subsection{Information Ablation}
\label{app:ablation}
The three inputs differ only in the text appended after the question; the visual input is identical. $V{+}L$ appends the log of subtasks completed before the decision point. $V{+}L{+}E$ appends the same log followed by the \texttt{accident\_cause} and \texttt{accident\_consequence} of completed anomalous subtasks. Table~\ref{tab:app_ablation_input} shows an example for one question.

\begin{table}[H]
\centering
\caption{Appended input for one question (\texttt{Long-Horizon\_L2\_task3\_q1}; excerpt of subtask 18).}
\label{tab:app_ablation_input}
\small
\begin{tabularx}{\linewidth}{@{}lY@{}}
\toprule
Condition & Appended input (excerpt) \\
\midrule
$V{+}L$ & \texttt{\{"subtask": 18, "action": "The arm hits the clean apple on the table", "start\_sec": 80.6, "stop\_sec": 82.067, "left\_hand": false, "right\_hand": false, "navigation": false\}} \\
\addlinespace
$V{+}L{+}E$ & The same log, plus \texttt{\{"subtask": 18, "action": "The arm hits the clean apple on the table", "accident\_cause": "The arm touches a clean apple on the tabletop.", "accident\_consequence": "The apple rolled onto the ground."\}} \\
\bottomrule
\end{tabularx}
\end{table}

The actual input contains all logs completed before the decision point. $\Delta L=(V{+}L)-V$ and $\Delta E=(V{+}L{+}E)-(V{+}L)$ are both computed on the same questions (Table~\ref{tab:app_ablation}).

\begin{table}[H]
\centering
\caption{$P$ accuracy under the three inputs and paired gains.}
\label{tab:app_ablation}
\small
\setlength{\tabcolsep}{4pt}
\begin{tabular}{lrrrrr}
\toprule
Model & $V$ (\%) & $V{+}L$ (\%) & $V{+}L{+}E$ (\%) & $\Delta L$ (pp, 95\% CI) & $\Delta E$ (pp, 95\% CI) \\
\midrule
Gemini 3.8 Flash & 49.7 & 53.0 & 72.8 & $+3.3$ [$-4.6$, $+11.3$] & $+19.9$ [$+12.9$, $+27.2$] \\
Qwen3.5 397B-A17B & 37.7 & 36.4 & 53.6 & $-1.3$ [$-7.2$, $+4.3$] & $+17.2$ [$+10.2$, $+24.5$] \\
Qwen3.8 Omni Flash & 35.8 & 38.4 & 55.6 & $+2.6$ [$-4.0$, $+9.4$] & $+17.2$ [$+11.0$, $+23.7$] \\
Qwen3.8 27B & 31.8 & 33.1 & 43.0 & $+1.3$ [$-4.2$, $+7.0$] & $+9.9$ [$+3.9$, $+16.1$] \\
Qwen3.8 Flash & 31.1 & 32.5 & 42.4 & $+1.3$ [$-4.7$, $+7.6$] & $+9.9$ [$+2.7$, $+17.2$] \\
Qwen3.5 122B-A10B & 30.5 & 33.1 & 37.1 & $+2.6$ [$-2.1$, $+7.4$] & $+4.0$ [$-1.2$, $+9.3$] \\
\bottomrule
\end{tabular}
\end{table}

The decision types in Figure~\ref{fig:exp_analysis}(d) are determined by the stage transition from the completed subtasks to the correct next step:
\begin{itemize}
    \item \textbf{Change plan} (71 questions): entering recovery, exiting recovery, or responding to an anomaly that has just occurred, including a new anomaly during recovery.
    \item \textbf{Keep plan} (80 questions): continuing along the main task line, or remaining within a recovery process.
\end{itemize}

Every question falls into one of the two types. The annotation gain for change-plan decisions exceeds that for keep-plan decisions by $+10.4$ points ([$+2.4$, $+18.4$]).

\subsection{Joint Analysis}
\label{app:joint}
\paragraph{Independence expectation.} If correctness on $P$ and $C$ were independent, the expected proportion of pairs with both answers correct would be the product of the marginal accuracies: $33.9\% \times 48.7\% = 16.5\%$ (95\% CI [13.4, 19.7]). The observed proportion is 16.3\% ([12.7, 20.2]), differing from the expectation by $-0.2$ points ([$-1.9$, $+1.6$]).

\paragraph{Model-adjusted difference.} A linear probability model is fitted to the 2,416 paired responses, with $P$ correctness as the dependent variable and $C$ correctness and model fixed effects as regressors. The coefficient on $C$ is $-2.7$ points ([$-10.04$, $+4.73$]); without controlling for model, the difference is $-0.6$ points ([$-7.60$, $+6.43$]). Adding question-pair fixed effects, which absorb item difficulty, gives $-3.5$ points ([$-7.57$, $+0.58$]; 2,000 task-clustered replicates).

\subsection{Causal CoT}
\label{app:cot}
Direct answering and causal CoT are evaluated in the same run on the same 151 questions; the prompts are given in \hyperref[app:prompts]{Appendix~\ref*{app:prompts}}. The direct accuracy of Gemini 3.8 Flash in this run (47.0\%) differs slightly from its main-evaluation $V$ accuracy on the same questions (49.7\%, Table~\ref{tab:app_ablation}), as its API is nondeterministic even at temperature 0, whereas the two Qwen models reproduce their main-evaluation results; Tables~\ref{tab:exp_cot} and~\ref{tab:app_cot} use the same-run results throughout. The $C$ groups follow the $C$ results of the main evaluation. In Table~\ref{tab:app_cot}, \emph{corrected} means incorrect under direct answering and correct under CoT, and \emph{newly wrong} means correct under direct answering and incorrect under CoT. The net gain is $100 \times (\text{corrected} - \text{newly wrong}) / \text{number of questions}$.

\begin{table}[H]
\centering
\caption{Answer changes under causal CoT.}
\label{tab:app_cot}
\small
\setlength{\tabcolsep}{4pt}
\begin{tabular}{lrrrr}
\toprule
 & \multicolumn{2}{c}{All 151 questions} & \multicolumn{2}{c}{$C$-correct group} \\
\cmidrule(lr){2-3}\cmidrule(lr){4-5}
Model & Corrected / newly wrong & Net gain (pp, 95\% CI) & Questions & Corrected / newly wrong \\
\midrule
Gemini 3.8 Flash & 21 / 9 & $+7.9$ [$+1.4$, $+14.7$] & 108 & 19 / 5 \\
Qwen3.8 Flash & 18 / 13 & $+3.3$ [$-3.6$, $+10.6$] & 67 & 7 / 4 \\
Qwen3.8 27B & 13 / 19 & $-4.0$ [$-12.1$, $+3.6$] & 65 & 8 / 8 \\
\bottomrule
\end{tabular}
\end{table}

We also repeat the joint analysis of Section~\ref{sec:joint_analysis} under both prompts. The difference in $P$ accuracy between correct-$C$ and incorrect-$C$ pairs changes from $-12.3$ ([$-30.8$, $+7.0$]) to $+5.3$ ([$-13.2$, $+23.2$]) points for Gemini 3.8 Flash (change $+17.6$, [$+5.6$, $+31.1$]), from $-5.0$ to $-2.9$ for Qwen3.8 Flash (change $+2.1$, [$-12.5$, $+16.4$]), and from $-1.8$ to $+5.2$ for Qwen3.8 27B (change $+7.0$, [$-6.7$, $+22.3$]). Intervals use the task-clustered bootstrap with 2,000 replicates.

Table~\ref{tab:app_cot_cases} shows two wrong-to-right responses per model, one for a self-induced anomaly (L2) and one for an external intervention (L3). In the selected examples, the CoT description of the event agrees with the annotation. Model outputs are reproduced verbatim; the option text of each final answer is given in parentheses.

\begin{table}[H]
\centering
\caption{Wrong-to-right examples under causal CoT.}
\label{tab:app_cot_cases}
\footnotesize
\setlength{\tabcolsep}{4pt}
\renewcommand{\arraystretch}{1.15}
\begin{tabularx}{\linewidth}{@{}>{\raggedright\arraybackslash}p{0.11\linewidth} >{\raggedright\arraybackslash}p{0.2\linewidth} >{\raggedright\arraybackslash}p{0.16\linewidth} Y@{}}
\toprule
Model & Task and key event & Direct answer (incorrect) & CoT output (correct) \\
\midrule
Gemini 3.8 Flash & Wipe a tablet; the wipe packet slips while being torn open and the wipes scatter on the floor (L2). & Move over to the front of the table. & ``Assessment: When opening the packet of wipes, it tore aggressively and the wipes scattered onto the floor by the shelf. The user just discarded the wrapper into the trash can and now needs to collect the wipes. \newline Final Answer: [B]'' (Go over to where the wipes are scattered.) \\
\addlinespace
Gemini 3.8 Flash & Prepare a fruit plate; another person takes some grapes, leaving fewer than 12 (L3). & Carry the fruit plate onward into the living room. & ``Assessment: Another person took grapes from the plate while the user was wiping their hands, leaving fewer than the required 12 grapes. More grapes must be added to fulfill the instruction. \newline Final Answer: D'' (Add grapes onto the fruit plate to fill it.) \\
\addlinespace
Qwen3.8 Flash & Spray anti-mite spray on bedding; the robot arm knocks over the balcony trash can and garbage scatters (L2). & Carry on straightening the quilt out for drying. & ``Assessment: The task of spraying the bedding is complete, and the person has moved to hang the quilt on a rack. A black trash can has fallen over with garbage scattered on the floor, which is an obstacle in the current path that needs to be addressed before continuing. \newline Final Answer: C'' (Step over to where the trash can fell.) \\
\addlinespace
Qwen3.8 Flash & Boil a kettle of water; the kettle is switched off and moved onto the counter before the water boils (L3). & Raise the kettle lid without lifting the kettle. & ``Assessment: The kettle was just moved off the heating base to the right side of the counter, so it is no longer positioned for boiling. To resume the task, the next step is to pick up the kettle from its current location on the right. \newline Final Answer: C'' (Pick up the kettle from the counter's right.) \\
\addlinespace
Qwen3.8 27B & Get a bottle of soda; the bottle had fallen earlier, and foam overflows after the cap is loosened (L2). & Turn the bottle cap to release trapped gas. & ``Assessment: The bottle was opened and is now foaming over the top, with foam visible on the neck. The most logical next step to clean up this mess before proceeding is to wipe the foam off the bottle and counter. \newline Final Answer: C'' (Wipe the foam off the bottle and counter.) \\
\addlinespace
Qwen3.8 27B & Clear standing water in a bathroom; another person walks in and out of the shower area, leaving wet footprints on the mopped floor (L3). & Mop the damp patch beside the bathroom cabinet. & ``Assessment: A second person entered the bathroom and left new wet footprints on the floor, which disrupted the cleaning. The most logical next step is to mop up these new wet footprints one by one. \newline Final Answer: B'' (Mop up the new wet footprints one by one.) \\
\bottomrule
\end{tabularx}
\end{table}

\subsection{Human Evaluation}
\label{app:human}
Two evaluators holding at least a bachelor's degree independently answered all 699 questions (548 $P$ and 151 $C$).
\begin{itemize}
    \item \textbf{Presentation.} As in the model input, each question presents the history video, the current decision frame, the task goal, the question, and four options in order. For $C$ questions, the history video carries the same \texttt{OPTION A}--\texttt{D} marks, and the question states the next action.
    \item \textbf{Answering.} After viewing, the evaluators select one option for each question. They do not discuss the questions with each other and have no access to the action annotations, causal annotations, or reference answers.
    \item \textbf{Scoring.} Scoring is the same as for the models. The Human row in Table~\ref{tab:main} averages the accuracies of the two evaluators and is excluded from ranking.
\end{itemize}
Both evaluators answered all $C$ questions correctly; their $P$ errors fall in L2--L4, with none in L1 (Table~\ref{tab:app_human}). The number of questions per level is 205, 113, 128, and 102 for $P$ (L1--L4) and 58, 60, and 33 for $C$ (L2--L4).

\begin{table}[H]
\centering
\caption{Human evaluation results.}
\label{tab:app_human}
\small
\setlength{\tabcolsep}{4pt}
\begin{tabular}{lcrrrrrcrr}
\toprule
 & & \multicolumn{5}{c}{$P$ accuracy (\%)} & & & \\
\cmidrule(lr){3-7}
Evaluator & $P$ errors (L1/L2/L3/L4) & L1 & L2 & L3 & L4 & All & $C$ errors & $C$ (\%) & Overall (\%) \\
\midrule
Evaluator 1 & 13 (0/4/3/6) & 100.0 & 96.5 & 97.7 & 94.1 & 97.6 & 0 & 100.0 & 98.1 \\
Evaluator 2 & 11 (0/3/3/5) & 100.0 & 97.3 & 97.7 & 95.1 & 98.0 & 0 & 100.0 & 98.4 \\
\midrule
Mean & 12 & 100.0 & 96.9 & 97.7 & 94.6 & 97.8 & 0 & 100.0 & 98.3 \\
\bottomrule
\end{tabular}
\end{table}

\section{Limitations}
\label{app:limitations}
While EMBER-Bench covers four household scenario types of cross-event causal memory, there are still additions from which it would benefit. For example, as we continue data collection, we will expand beyond the current 189 tasks and 699 questions to more diverse household scenes, and extend the evaluation from offline multiple-choice questions on pre-recorded videos toward closed-loop interaction. Additionally, the benchmark currently provides test data only; constructing companion training resources that are disjoint from the test set would further support research on memory mechanisms and fine-tuning. Finally, the causal CoT in Section~\ref{sec:causal_reasoning} is a basic probe evaluated on three models, and it is unclear to what extent more refined reasoning strategies would close the cause--action gap. We plan to explore such strategies, validate them across a wider range of models, and continue updating the benchmark as new models are released.

\end{document}